\documentclass[runningheads]{llncs}

\usepackage{eccv}

\usepackage{eccvabbrv}
\usepackage{listings}
\usepackage{graphicx}
\usepackage{booktabs}

\usepackage[accsupp]{axessibility}  

\usepackage{hyperref}

\usepackage{orcidlink}

\usepackage{multirow}
\usepackage{adjustbox}

\newcommand{\revise}[1]{{\color{blue}#1}}

\begin{document}

\newcommand{\DiffPMAE}{\textit{DiffPMAE{}}}
\title{DiffPMAE: Diffusion Masked Autoencoders for Point Cloud Reconstruction} 

\titlerunning{DiffPMAE}

\author{Yanlong Li\inst{1}
\and
Chamara Madarasingha\inst{2} \and
Kanchana Thilakarathna\inst{1}
}

\authorrunning{Y.~Li et al.}

\institute{The University of Sydney\\
\email{yali8838@uni.sydney.edu.au} \email{kanchana.thilakarathna@sydney.edu.au}\\
\and
University of New South Wales\\
\email{c.kattadige@unsw.edu.au}}

\maketitle

\begin{abstract}
  Point cloud streaming is increasingly getting popular, evolving into the norm for interactive service delivery and the future Metaverse. However, the substantial volume of data associated with point clouds presents numerous challenges, particularly in terms of high bandwidth consumption and large storage capacity. Despite various solutions proposed thus far, with a focus on point cloud compression, upsampling, and completion, these reconstruction-related methods continue to fall short in delivering high fidelity point cloud output. As a solution, in \DiffPMAE{}, we propose an effective point cloud reconstruction architecture. Inspired by self-supervised learning concepts, we combine Masked Autoencoder and Diffusion Model to remotely reconstruct point cloud data. By the nature of this reconstruction process, \DiffPMAE{} can be extended to many related downstream tasks including point cloud compression, upsampling and completion. Leveraging ShapeNet-55 and ModelNet datasets with over 60000 objects, we validate the performance of \DiffPMAE{} exceeding many state-of-the-art methods in terms of autoencoding and downstream tasks considered. Our source code is available at : \href{https://github.com/TyraelDLee/DiffPMAE}{https://github.com/TyraelDLee/DiffPMAE}
  \keywords{Point Cloud compression \and Masked Autoencoder \and Diffusion model}
\end{abstract}

\section{Introduction}

\begin{figure}[t]
  \centering
   \includegraphics[width=0.8\linewidth]{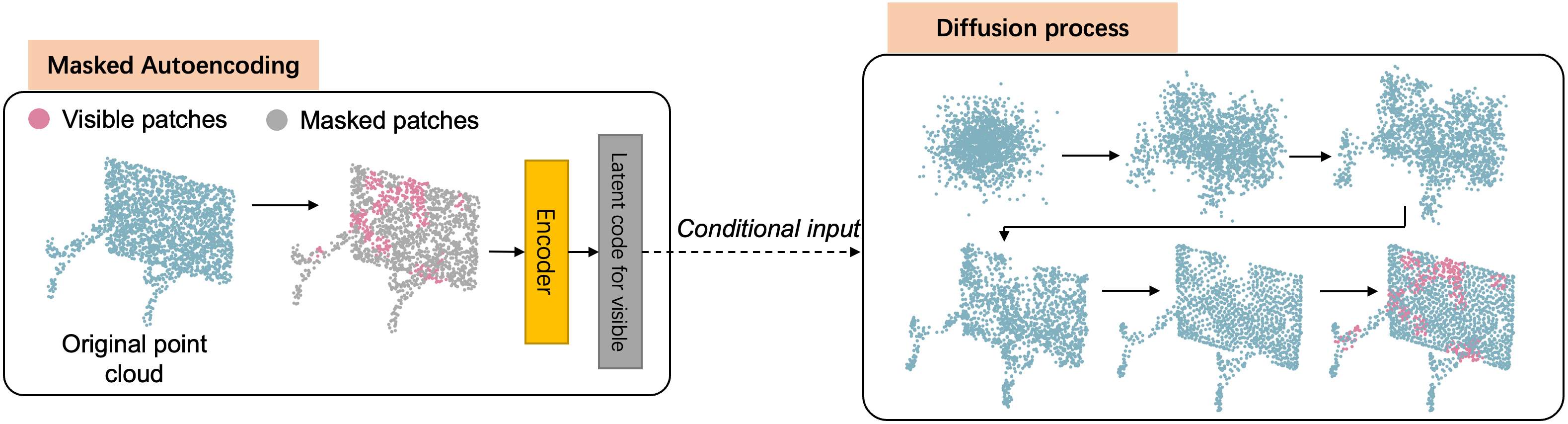}
   
   \caption{\DiffPMAE{} inference process: MAE module first segments the point cloud to visible and masked regions and provides latent code for visible patches which is taken as a conditional input for the Diffusion process.   
   DM reconstruct masked regions from noise which is combined with visible patches.}
   \label{fig:teaser}
\end{figure}

Advancements in immersive multimedia technology, in particular the widespread availability of devices for generating and consuming 3D visual data, hold tremendous potential for realizing interactive applications and the future Metaverse. In light of that, point clouds have emerged as a prominent format for volumetric data facilitating 3D vision. As the use of point clouds continues to soar, various challenges have already been posed against providing high quality 3D content primarily due to its complexity which can grow to millions of points bearing multiple attributes~\cite{chen2023introduction}. For example, efficient \emph{storage and transmission} of point cloud has recently become a pressing concern~\cite{nardo2022point, shi2023enabling, quach2022survey} that are exacerbated by inevitable challenges such as data losses in network, limited storage and bandwidth with other competing applications, etc.

There are many solutions proposed to date to address these issues such as 
point cloud compression algorithms including both statistical (e.g., octree, \cite{octree_1, octree_2, octree_3}, kd-tree \cite{kdtree_1, kdtree_2, draco}) and deep learning~\cite{dpcc, Song_2023_CVPR, compression, compression_ai_1}) methods followed by upsampling~\cite{pugan, putrans, 3pu, dispu} to compress the data and reconstructing them before the user consumption. Though these methods can reduce the bandwidth and storage consumption, they still lack high fidelity reconstruction at high compression ratio. In the meantime, different point cloud completion solutions are proposed to reconstruct point cloud objects with limited available data~\cite{Snowflake, pcn, pointr}. However, they often fail to generate accurate output, particularly, if the missing regions of the point cloud are significant and highly random. Hence, there is a growing demand for efficient models for point cloud \textit{reconstruction}, particularly when the available data for a given point cloud object is limited involving higher levels of \textit{compression} or data losses.

On the other hand, self-supervised learning (SSL) models have demonstrated outstanding performance in natural language processing (NLP) through exceptional models such as GPT \cite{GPT}, BERT \cite{BERT} and other models \cite{UnsupervisedNLP_1, UnsupervisedNLP_2, UnsupervisedNLP_3}. SSL for point clouds has also been proposed focusing particularly on generating missing point cloud data for a given object, e.g., PointMAE \cite{PointMAE}, PointM2AE~\cite{pointm2ae}, and \cite{CrossPoint, pcssl_1, pcssl_2, pcssl_3}. 
However, these approaches require both encoder and decoder to be trained together limiting its applicability in point cloud transmission tasks requiring prior collaboration between client and server. In the meantime, they do not thoroughly evaluate on the applicability of SSL for point cloud reconstructions and the related downstream tasks (e.g., compression, upsampling).



To this end, we propose \DiffPMAE{}, a deep learning based self-supervised point cloud \
reconstruction architecture that combines Masked Autoencoder (MAE) with Diffusion Models (DM). 
Our model segments the point cloud data to masked-visible patches and then takes latent space of visible patches as the condition to guide masked token generation (retrieved by our decoder) in the diffusion process as illustrated in Fig.~\ref{fig:teaser}. In the meantime, the encoder and decoder in MAE can be trained separately providing more flexibility in model training and distributing them in streaming applications.
Thus, by the nature of the reconstruction process, the proposed model can easily be extended to point cloud processing tasks including compression, completion and upsampling. Due to its unstructured nature, there is no meaningful geometrical relation between visible patches of the input point cloud and Gaussian noise which is used as the conventional learning objective in DMs. We overcome these challenges in developing a robust architecture to effectively leverage the capabilities of Diffusion process. 

With comprehensive empirical experiments with pre-training on ShapeNet-55 dataset and ModelNet validation sets, \DiffPMAE{} outperforms state-of-the-art generative models. In the autoencoding performance, \DiffPMAE{} achieves  $21.9\%$ of average improvement in Minimum Matching Distance Chamfer Distance (MMD CD) 
compared with benchmarks. 
Considering the downstream point cloud related tasks, \DiffPMAE{} outperforms many other state-of-the-art methods. For example,
in upsampling task, \DiffPMAE{} outperforms recent works by 31\% improvements on average in MMD CD.  \DiffPMAE{} also provides a competitive compression ratio along with an average improvement of 67.7\% in decompression quality and 73.2\%  average improvement in point cloud completion tasks in the related metrics.



A summary of our main contributions include:
\begin{itemize}
    
    \item We introduce \DiffPMAE{}, a self-supervised learning based model which combines Diffusion Models and MAE models for {point cloud  reconstruction}.
    \item We demonstrate that the proposed MAE modules are strong in creating non-trivial latent representations of limited point cloud data that can be used as conditional input for DMs to reconstruct missing point cloud regions. 
    \item We extend \DiffPMAE{} for various point cloud processing tasks including point cloud compression, upsampling and completion in which we outperform various state-of-the-art approaches.
    \item We extensively validate the \DiffPMAE{} by a thorough ablation study highlighting the significance of design strategies taken.

  
\end{itemize}

\section{Related Work}
\label{sec:related}

\noindent
\textbf{Self-supervised learning} (SSL) aims to learn from the information or structure of data itself instead of human-labelled data. SSL has been well developed in NLP, such as GPT \cite{GPT} and BERT \cite{BERT}. 
For point cloud tasks, SSL have also been researched comprehensively \cite{PointMAE, pcssl_1, pcssl_2, pcssl_3, pcssl_4, CrossPoint}. For example, PointBERT \cite{pointbert} proposed a BERT-styled approach that segmented and masked input tokens to predict the masked parts via discrete Variational Autoencoder (VAE). Denoising Autoencoder (DAE) \cite{dae_1, dae_2} is a type of autoencoder that aims to enhance the performance by adding noises to inputs. By extending the idea from DAE, the MAE \cite{MAE} replaced the input noise with masked patches. A recent work, DiffMAE \cite{DiffMAE}, proposed an approach that replaced the decoder in the MAE with a diffusion model for the image related works to achieve better results. 
In computer vision tasks, PointMAE \cite{PointMAE} bring the masked autoencoder idea to point cloud tasks by randomly mask input patches and make predictions on them.


\noindent
\textbf{Point cloud related tasks:} We identify three main tasks related with point cloud data that are directly comparable with \DiffPMAE{} namely, point cloud \textit{compression}, \textit{completion} and \textit{upsampling}. Among statistical approaches for point cloud \textit{compression},  octree~\cite{octree_1, octree_2, octree_3} and kd-tree-based compression \cite{kdtree_1, kdtree_2, draco} are the most prominent methods, which follows hierarchical compressing approach by voxelizing the point cloud data.  DNN architectures are also proposed to convert point clouds to low-dimensional latents~\cite{dpcc, compression, compression_ai_1}. 
In \textit{upsampling}, related research has evolved from CNN methods like PU-Net \cite{punet} to sophisticated generative models such as GAN and transformers \cite{pugan, putrans} while increasing the upsampling performance. 
In point cloud \textit{completion}, the folding-based autoencoder~\cite{foldingnet} proposed a two-step decoding that combined the 2D grid with the 3D point cloud data. The coarse to fine autoencoding \cite{pfnet, cascaded, pcn} has introduced the two stages of completion, which generates the coarse results first and then increase the density of coarse low-resolution point clouds.

\noindent
\textbf{Diffusion models (DM)}
are widely studied in 1D and 2D domains~\cite{hrConditionIG_1, hrConditionIG_2, hrConditionIG_3, ddpm,tti_1, tti_2}, however, is  still a new concept for  3D point cloud generation.
The  Luo \textit{et al.} \cite{pointDiffusion} proposed an autoencoder architecture with a DM as a decoder. The work in \cite{pointe} proposes a diffusion-based text and image-conditioned point cloud generation, but the performance is still insufficient due to the complexity of the 3D data. LION \cite{lion} directly works on the latent space to generate 3D shapes. The work \cite{gecco} proposed an image-conditioned point cloud generative DM with the geometrical information to improve the quality.

\textit{
In contrast, \DiffPMAE proposes a SSL based approach which leverage DMs combining with MAE \cite{PointMAE} with transformer architectures for point cloud generation. We further show how it can be extended for various downstream tasks including point completion, compression and upsampling.}



\section{Methodology}
\label{sec:method}
\begin{figure*}
  \centering
   \includegraphics[width=0.9\linewidth]{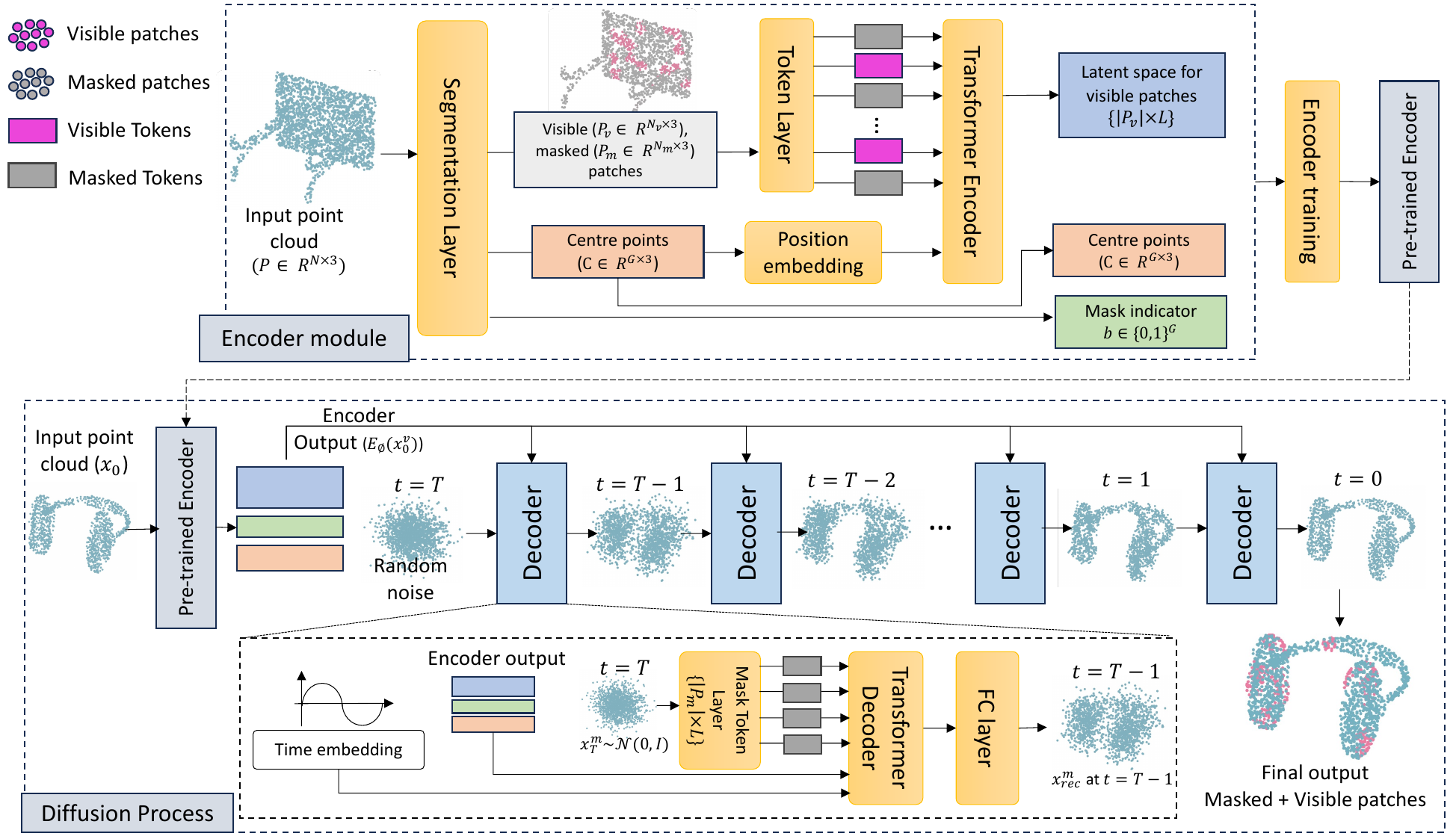}
   \caption{Overall structure for \DiffPMAE{} containing the MAE and DM module. During training, the encoder module will be trained first. The pre-trained encoder will be used to encode the point cloud input to latent code for diffusion model training.}

   \label{fig:overview}
\end{figure*}
\subsection{Overview of \DiffPMAE}
\label{subsec:overview}

Fig.~\ref{fig:overview} illustrates the end-to-end process of \DiffPMAE. As the first step, we pre-train an Encoder module $E_{\phi}$ which first segments original point cloud object into visible and masked patches. Then, these regions are passed through a Token layer that generates a tokenized output for the patches before feeding them to a Transformer encoder. 

Next, we train a DM which iteratively adds noise in the forward process until the input point cloud object with only masked patches becomes completely Gaussian noise. Then, we perform a denoising process starting from Gaussian random noise to the expected point cloud in $T$ steps.


\subsection{Encoder module}\label{subsec:encoder}

\noindent
\textbf{Segmentation layer.}~ 
We denote the input point cloud as $P \in \mathbb{R}^{N \times 3}$ with three dimensional ($x,y,z$ coordinates) $N$ number of points. 
We first use the FPS (Farthest Point Sampling) algorithm to determine centre points for $G$ number of groups, and then we use the K-nearest neighbourhood (K-NN) algorithm to find points for each corresponding group. After segmenting, we randomly split the segmented patches into masked patches $P_m \in \mathbb{R}^{N_m \times 3}$ and visible patches $P_v \in \mathbb{R}^{N_v \times 3}$, where $N_m$ and $N_v$ are number of points in respective segments. We denote the mask ratio $r$ as the proportion of masked points. The segmentation layer generates two other outputs vectors: 
(i) centre points $C \in \mathbb{R}^{G \times 3}$ for each patch, and 
(ii) a binary indicator $b \in \{0,1\}^G$ whether the patch is masked or not as the meta data for decoding. 
We feed centre points to a simple NN layer to receive position embedding that will be an input to the Transformer encoder.

\noindent
\textbf{Token layer.}~$P_m$ and $P_v$ patches from the Segmentation layer are fed to a token layer to convert them to a latent code before a transformer module. Experimentally we observed that such latent output which is rich in extracted patch features can enhance the Transformer performance. Here, the Token layer is a simple convolutional neural network module providing $L$ long 1D latent code for each patch in $P_v$ and $P_m$.

\noindent
\textbf{Transformer.}~ We use a standard transformer but without a decoder~\cite{attention}. We input the latent space from the token layer and the corresponding position embedding to this transformer to generate the latent code of the visible patches. Here, the position embedding are the encoded representation of centre points from segmentation layer through simple linear NN layer. In order to control the range of latent code, we add normalization layer after the transformer. 


\noindent
\textbf{Training the Encoder.}\label{subsubsec:train encoder}
To pre-train our Encoder separately from the Decoder in the diffusion process (see Section~\ref{subsec:diffusion_process}), we apply an extra fully connected layer (FC layer) directly after the Transformer. That layer will convert encoded latent to point clouds that can be used to calculate the loss compared with the ground truth.
Here, we select Chamfer Distance-$L2$ as the loss function~\cite{cdl2} as in Eq.~\ref{eq:encoder loss}. 

\begin{equation}
  \mathcal{L}_{sampling} = \text{CD L2}(E_{\phi}(P), P)
  \label{eq:encoder loss}
\end{equation}

\noindent
where  $E_{\phi}(P)$ is the output of our Encoder module (after the fully connected layer) and  $P$ is the ground truth point cloud. All the trainable parameters in the above components are trained together. 

As we do not add decoder parts during this training,  
the speed of the training can be accelerated. Note that
by splitting the Encoder and the Decoder, we expect 2 main advantages. First
our encoder model can be fine-tuned for performance improvement or other downstream tasks, such as upsampling, agilely by training the Encoder or the Decoder independently. Second, pre-trained encoder provides stable conditional inputs to the DM process.
The trained Encoder has three main vector outputs namely,
\textit{i})~Encoded visible latent with size $\{|P_v|, L\}$ where $|P_v|$ is the size of visible patches, $L$ is the latent length, \textit{ii})~Centre points $C$ and \textit{iii})~Binary mask indicator from Segmentation layer as shown in Fig.~\ref{fig:overview}.


\subsection{Diffusion process}
\label{subsec:diffusion_process}

We train a DM as the Decoder of \DiffPMAE which takes the Encoder output as the conditional input for DM and the output from the Encoder for each point cloud model is fixed. The backbone network for predicting the reverse diffusion process is an encoder-only Transformer combined with other key modules below. 

\noindent
\textbf{Time embedding.}~This layer 
provides a unique embedding for each time step in the diffusion sequence 
which allows the Decoder transformer to learn the temporal relation and handle the time sequence. At each time step $t$, we extract a corresponding sinusoidal position which is then converted to a time embedding through a simple NN layer.

\noindent
\textbf{Mask Token layer.}~This layer 
converts different noise distributions in the diffusion process to a latent space
while controlling the number of masked patches according to the binary mask indicator from the Encoder. This helps to keep a constant number of samples in the output and perfectly aligns the masked and visible patches, which has a noticeable impact on the overall performance of the DM. 
The size of input noise distribution for the Mask Token layer is ${N_m \times 3}$, where $N_m$ is the number of masked points. By default, our model processes 2048-sized input point cloud (i.e., $P_m + P_v$) with $r=0.75$ mask ratio resulting in $P_m=1536$.  The size of output from the mask token layer is $\{|P_m| \times L\}$.

\noindent
\textbf{Transformer.}~We follow a transformer architecture in the Decoder similar to that in the Encoder.
However, in the Decoder transformer, we add normalization layers between each block to control the range of the intermediate outputs from the block and accelerate the convergence speed. The input of this Transformer is a $\{G \times L\}$ vector. That input vector concatenates the masked latent from the Mask token layer and the visible latent from the Encoder. The output from this Transformer is a predicted masked latent. Additionally, we connect an FC layer to the last block in the transformer, which converts the output from the transformer to a point cloud object.

\subsection{Training the diffusion model.}
\label{subsubsec:train_diffusion}


Given an original training sample $x_0\sim p(x_0)$, the masked and visible samples are denoted by $x_0^m \in P_m$ and $x^v_0 \in P_v$ respectively. The main task of our diffusion process is to sample $x^m_0$ conditioned by $E(x^v_0)$, which is the encoded visible patches received from the pre-trained encoder. In the forward pass, we recursively add a small amount of Gaussian noise on $x^m_0$ until $T$ times creating a sequence of $x^m_0$,$x^m_1$,$x^m_2$, ..., $x^m_T$. We treat this process as a Markov process as in Eq.~\ref{eq:markov}

\begin{equation}
  q(x^m_t|x^m_{t-1}) = \mathcal{N}(x^m_t;\sqrt{1-\beta_t}x^m_{t-1},\beta_t\mathbf{I})
  \label{eq:markov}
\end{equation}

\noindent
where $t\in[1,2,3,...,T]$ is the timesteps and $\beta_t\mathbf{I}$ is the variance of the noise at $t$. Following the properties of a Gaussian distribution we can convert Eq.~\ref{eq:markov} to Eq.~\ref{eq:markov-re} by removing the recursion for simplified operation by denoting $\alpha_t=1-\beta_t$ and $\overline{\alpha}_t=\prod_{i=1}^t\alpha_i$

\begin{equation}
  q(x^m_t|x^m_0) = \mathcal{N}(x^m_t;\sqrt{\overline{\alpha}_t}x^m_{t-1},(1-\overline{\alpha}_t)\mathbf{I}) 
  \label{eq:markov-re}
\end{equation}

A reparameterization can convert Eq.~\ref{eq:markov-re} to $\sqrt{\overline{\alpha}_t}x^m_0~+~\sqrt{1-\overline{\alpha}_t}\epsilon$ where $\epsilon\sim \mathcal{N}(\textbf{0},\textbf{I})$. With a small $\beta$ at each step, we can assure  $x^m_T\sim \mathcal{N}(\textbf{0},\textbf{I})$. 

In the reverse process, we aim to get distribution $q(x^m_{t-1}|x^m_t, E(x^v_0) )$ step by step to recover the input from $x^m_T\sim \mathcal{N}(\textbf{0},\textbf{I})$.  
However, sampling from $q(x^m_{t-1}|x^m_t, E(x^v_0) )$ is not an easy task without knowing the entire diffusion process. 
Therefore, we train the Decoder module with transformers to learn  $p(x^m_{t-1}|x^m_t, E(x^v_0))$ to get the conditional probabilities approximately to infer the entire reverse diffusion process. Unlike standard DM, which predicts Gaussian noise as the objective, \DiffPMAE{} uses $x^m_0$ as the objective. This is because standard Gaussian noise does not maintain any geometrical relationship with visible features. To get the $x^m_0$ during sampling, we train a deep learning model to learn the distribution of $x^m_0$ as $x^m_{rec}$ at timestep $t$ with the condition $E(x^v_0)$. Here, $x^m_{rec}$ is the output from the Decoder at each time step. 

As a result, the sampling process can be presented as below, where $\sigma_t = 1-\overline{\alpha}_t$.

\begin{equation}
    \begin{aligned}
      q(x^m_{t-1}&|x^m_t,x^m_0,E(x^v_0)) = \\ 
      &(\frac{\sqrt{\alpha_{t}} (1-\overline{\alpha}_{t-1})}{1 - \overline{\alpha}_{t}}x^m_{t} + \frac{\sqrt{\overline{\alpha}_{t-1}}\beta_{t}}{1-\overline{\alpha}_{t}}x^m_{rec}) + \sigma_{t}x^m_{rec}
    \end{aligned}
  \label{eq:also-important}
\end{equation}


We extend the optimisation of the simple objective proposed by DDPM \cite{ddpm}. We use the Chamfer Distance-$L2$ as the loss function, which leads to better results for 3D point cloud tasks as in Eq.~\ref{eq:Loss DM}.

\begin{equation}
  \mathcal{L}_{sampling} = \text{CD L2}(D(x^m_{t}, t, E_\phi({x^v_0}), x^m_{0})
  \label{eq:Loss DM}
\end{equation}

\noindent
where $E_\phi$ is the Encoder function and $D$ denotes the Decoder module. 

We test two Decoder configurations that change the model output, \textit{i})~\textit{Config. 1}:~Predict only the $P_m$ from the DM. In this scenario, the mask token layer provides tokens to represent only the masked patches, \textit{ii})~\textit{Config. 2}:~Predict both $P_m$ and $P_v$, which is useful for upsampling task (see Section~\ref{upsampling}). In this configuration, tokens are generated to represent both masked and visible patches from the mask token layer.

\section{Experiments and Results}
\label{sec:experiments}
\subsection{Evaluation setup}\label{subsec:eval_setup}
\noindent
\textbf{Dataset.}~We use ShapeNet-55~\cite{shapenet2015}(55 categories, 52470 models) and ModelNet40~\cite{modelnet} (40 categories, 12311 models) for train and validation of the models. We downsampled the models to lower density to support upsampling tasks and to match the number of points in benchmark models from literature for a fair comparison. 
Each point cloud object is divided into 64 groups ($G$) in equal size. 

\noindent
\textbf{Important model details and hyper-parameters.}
The default  transformer encoder module contains 12 blocks. Each block includes 6 heads, and the latent width ($L$) is 384. The default transformer in the Decoder consists of 4 blocks with 4 headers in each block and the latent width is 384. The default mask strategy is randomly masking, and the default mask ratio ($r$)  is 75\%. The timestep ($T$) for the diffusion process is 200 steps, 
and the range of $\beta$ is $10^{-4}$ to $0.05$.
More details will be released with artefacts.

\noindent
\textbf{Evaluation metrics.}
We use 4 main metrics for the evaluation: 
\textit{i})~MMD CD (Minimum Matching Distance Chamfer Distance)~\cite{achlioptas2018learning}: matches every points between two objects for the minimum distance (MMD) and
report the average of distances in the matching using point-set distance,
\textit{ii})~1-NN CD (1-Nearest Neighbor Chamfer Distance)~\cite{pointDiffusion}: finds the nearest neighbour in the ground truth set using the K-NN algorithm, then calculates the distance between two point clouds. It reflects the overall generation performance,
\textit{iii})~JSD (Jensen-Shannon divergence)~\cite{jsd}:
assesses the similarity between the generated result sets and ground-truth sets on geometrical distribution,
and 
\textit{iv})~HD (Hausdorff Distance) \cite{hd}: computes the HD on each pair of generated results and ground truth and then calculates the average. 

\subsection{Autoencoding performance}
\begin{figure*}
  \centering
   \includegraphics[width=1\textwidth]{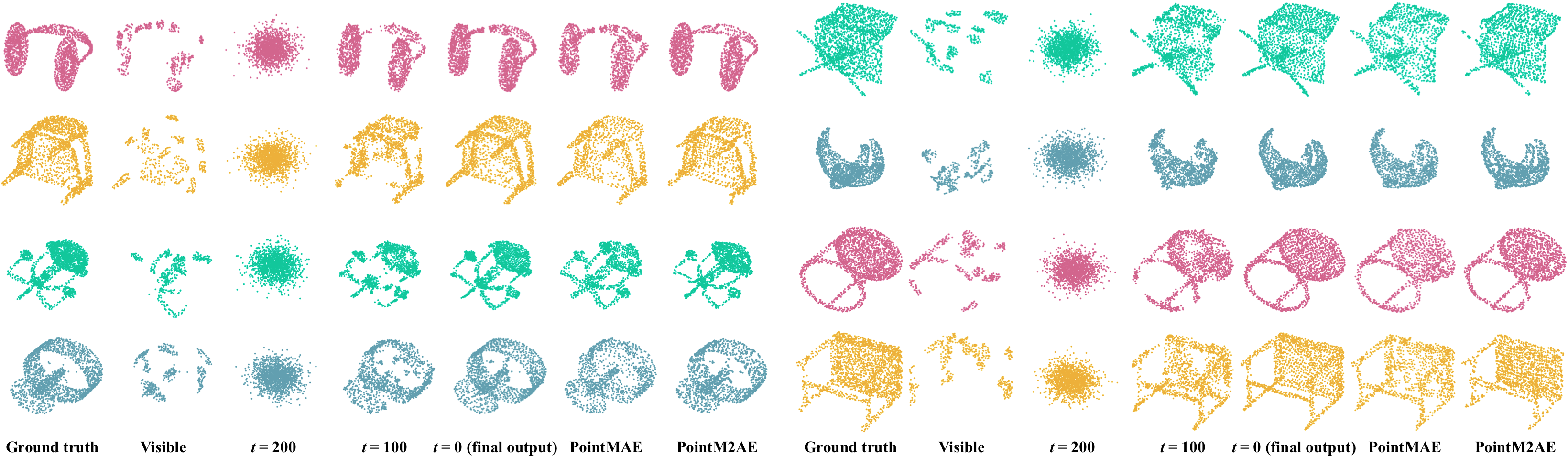}
   \caption{{Qualitative comparison of \DiffPMAE, PointMAE and PointM2AE.} $t=0$ is the final output from \DiffPMAE{} that combines visible parts and predicted masked parts. The $r$ for all methods is 75\%.
   }
   \label{fig:AE_compare}
\end{figure*}

\begin{table}[b]
    \small
    \centering
    \begin{adjustbox}{width=0.65\textwidth}
        \begin{tabular}{c|c|c|c}
        \hline
        Model & MMD CD ($\times10^{-3}$) & JSD ($\times10^{-3}$)  & HD ($\times10^{-2}$)\\
        \hline
        \hline
        PointMAE \cite{PointMAE} & 1.752 & 53.108 & 4.193 \\
        \hline
        PointM2AE \cite{pointm2ae} & 1.224 & \textbf{1.948} & 4.033 \\
        \hline
        Ours & \textbf{1.125} & 2.890 & \textbf{3.445} \\
        \hline
        \end{tabular}
    \end{adjustbox}
    \caption{Comparison of point cloud autoencoding performance. 
    }
    \label{tab:ae_compare_table}
\end{table}

We evaluate the autoencoding performance of \DiffPMAE{} comparing with benchmarks on ShapeNet-55 dataset.
The benchmarks we use are PointMAE \cite{PointMAE}, and PointM2AE \cite{pointm2ae} which are the most related and recent works comparable to the  MAE architecture in our method.  

Fig.~\ref{fig:AE_compare} visually compares the \DiffPMAE{} with  PointMAE, and PointM2AE. We notice that after 200 steps of reverse diffusion, our model can generate more uniform distributed point cloud samples compared to both benchmarks and keep local details in complex point cloud objects (chair: row 2 and table row-8) more precisely. {This is mainly due to the iterative sampling process introduced by DMs that can identify subtle feature variations in point cloud objects.} Table.~\ref{tab:ae_compare_table} reports a statistical summary of the comparison using MMD CD, JSD and HD metrics. We observe that \DiffPMAE{} outperforms all the methods in terms of MMD CD and HD values achieving 1.125 and 3.445 lowest values respectively. However, \DiffPMAE{} shows comparably lower performance in JSD mainly due to the higher uniformity in point cloud distribution which can be slightly deviated from ground truth data.


\subsection{Downstream tasks}
\begin{table*}[t]
    \small
    \centering
    \adjustbox{width=\columnwidth,center}{%
    \begin{tabular}{c|c|c|c|c|c|c|c}
    \hline
    Mask ratio ($r$) & Metrics & FoldingNet \cite{foldingnet} & PCN \cite{pcn} & PoinTr \cite{pointr}* & Snowflake \cite{Snowflake} & Ours (w/o pos)** & Ours (w/ pos)**\\
    \hline
    \hline
    \multirow{2}{*}{40\%} & MMD & 6.675 & 4.881 & - & 1.602 & 2.144 & \textbf{1.197} \\
    \cline{2-8}
                          & HD  & 9.467 & 8.388 & - & 3.534 & 3.494 & \textbf{2.586} \\
                          \hline
    \multirow{2}{*}{50\%} & MMD & 7.535 & 5.041 & - & 2.149 & 2.696 & \textbf{1.432} \\ 
    \cline{2-8}
                          & HD  & 9.987 & 8.535 & - & 3.919 & 4.295 & \textbf{3.125} \\
                          \hline
    \multirow{2}{*}{60\%} & MMD & 8.971 & 5.303 & - & 2.932 & 3.250 & \textbf{1.647} \\
    \cline{2-8}
                          & HD  & 10.789 & 8.773 & - & 4.411 & 5.022 & \textbf{3.598} \\
                          \hline
    \multirow{2}{*}{75\%} & MMD & 14.596 & 6.227 & 7.368 & 5.056 & 4.403 & \textbf{1.901} \\
    \cline{2-8}
                          & HD  & 13.419 & 9.537 & 6.515 & 5.697 & 6.363 & \textbf{4.193} \\
                          \hline
    \multirow{2}{*}{80\%} & MMD & 19.344 & 6.945 & - & 6.321 & 4.995 & \textbf{2.126} \\
    \cline{2-8}
                          & HD  & 15.288 & 10.063 & - & 6.246 & 6.874 & \textbf{4.650} \\     
    \hline
    \end{tabular}}
    \caption{Comparison of point cloud completion performance with two configuration of \DiffPMAE{}; with and without position embedding. MMD CD results are in $\times10^{-3}$, HD results are in $\times10^{-2}$. *The pre-trained model provided by PointTr only works at 75\% loss ratio in our experiments. **  The position embedding refers to the position for MAE, which differs from the position embedding in DMs.}
\label{tab:completate_table}
\end{table*}

\noindent
\textbf{Point cloud completion.}
Based on the nature of masking of the point cloud, \DiffPMAE{} can be used in point cloud completion tasks as well. To evaluate that, we fine-tune our Encoder model 
by removing the position embedding on the ShapeNet dataset and evaluating it on the ModelNet dataset. We compare~\DiffPMAE{} with existing point cloud completion models that are shown in Table~\ref{tab:completate_table} under two configurations; with (w/ pos) and without (w/o pos) position embedding, which is a conditional input from the encoder. 
For PoinTr, Snowflake and our model, we use point clouds with randomly missing content, which further increases the difficulty of the recovery task. We set the  $r$ to 40, 50, 60, 75 and 80\% reflecting different levels of point cloud loss. 

According to Table~\ref{tab:completate_table}, \DiffPMAE{} with position embedding achieves the best results in all different $r$ settings and is also noticeably better than without position embedding. Except for Snowflake, \DiffPMAE{} outperforms all other benchmarks. In Snowflake, we observe that MMD CD and HD scores are slightly better than \DiffPMAE{} without position embedding when the loss ratio is less than 75\%. 
{This is because Snowflake is originally designed for point cloud completion, therefore, it can achieve better performance when the $r$ is small.} 
However, the completion task is more like generation when the loss ratio is above 75\%. Hence, \DiffPMAE{} achieves better results with a high loss ratio showing its promise for completion tasks. NOTE: we also test on PCN dataset in the supplementary.

\begin{table}[b]
    \small
    \centering
    \adjustbox{width=\columnwidth,center}{%
    \begin{tabular}{c|c|c|c|c|c|c|c|c}
    \hline
    \multirow{2}{*}{Metrics} & PU-Net & 3PU & PU-GCN & Dis-PU & PU-Transformer & PU-Edgeformer & Ours  & Ours  \\
    & \cite{punet} & \cite{3pu} & \cite{pugcn} & \cite{dispu} & \cite{putrans} & \cite{kim2023pu} & (w/ 25\% Input) & (w/ 60\% Input) \\
    \hline  
    \hline
    MMD CD & \multirow{2}{*}{1.155} & \multirow{2}{*}{0.935} & \multirow{2}{*}{0.585} & \multirow{2}{*}{0.485} & \multirow{2}{*}{0.451} & \multirow{2}{*}{0.462} & \multirow{2}{*}{0.461} & \multirow{2}{*}{\textbf{0.423}} \\
    ($\times10^{-3}$) & & & & & & & & \\
    \hline
    HD & \multirow{2}{*}{15.170} & \multirow{2}{*}{13.327} & \multirow{2}{*}{7.577} & \multirow{2}{*}{4.620} & \multirow{2}{*}{3.843} & \multirow{2}{*}{\textbf{3.813}} & \multirow{2}{*}{5.130} & \multirow{2}{*}{4.487} \\
    ($\times10^{-3}$) & & & & & & & & \\
    \hline
    \end{tabular}}
    \caption{Comparison of point cloud upsampling performance.
    }
    \label{tab:upsample_table}
\end{table}

\noindent
\textbf{Upsampling.} 
\label{upsampling}
We run the DM in \textit{Config. 2} in which we sample both masked and visible patches together(see Section~\ref{subsec:diffusion_process}). 
We first train our pre-trained model for 100 epochs on ShapeNet for \textit{Config.~2} and then fine-tune the model on high-density ShapeNet for 50 epochs. During fine-tuning the model, we fixed the weight in the transformer and modified the structure of the prediction head. For the upsampling task, our model takes 2048 points model as input and generates 8192 points model.
We evaluate our model’s performance on PU1K dataset~\cite{pugcn}, which has been used by the benchmark models we compare \DiffPMAE{} with. 

Table~\ref{tab:upsample_table} reports upsampling performance measured by MMD CD score. We see that \DiffPMAE{} outperforms all other methods showing 
the lowest MMD CD value, 0.423. The main reason is \DiffPMAE{} effectively  learns the local features by using masking and the DM, which is more powerful in learning local features, can sample the points at a higher fidelity to original point cloud data. Moreover, our model only needs a part of the low density point cloud model (25\% - 60\% of number of points only) to achieve high-fidelity up-sampled output. \revise{}

\noindent
\textbf{Compression.}~In \DiffPMAE{} we masked $r$ ratio of point cloud data which indirectly implies that \DiffPMAE{} has the potential for  compressing  point cloud data. For example, with the default masking ratio, it is equivalent to compressing the point cloud data by 75\%. Streaming point cloud data is a direct use case of this downstream task where we can send only the visible patches from the Encoder to the client, rather than sending the entire point cloud data. 

As we experimentally observe that latent code for visible patches from the Encoder is relatively larger in size, we do not store or transmit this vector to avoid reducing compression gains. Alternatively, we propose to have a new Encoder which takes only the visible patches by changing the current Encoder (Fig.~\ref{fig:overview}) removing the segmentation layer. For example, in a point cloud streaming application, the client will have this trained Encoder so that the received visible data can be used to derive visible latent code without transmitting through the network. Note that we still use the current encoder, for example, at the content server according to the above streaming application, to derive centre points and masked binary vector required for DM.
We use the bpp (bits per point)~\cite{bbp} and mean Chamfer Distance~\cite{achlioptas2018learning} to evaluate the compression ratio and the decoded point cloud quality. We compare \DiffPMAE{} with existing point cloud compression approaches based on ShapeNet dataset.

Table~\ref{tab:bpp} shows that \DiffPMAE{} does not achieve the highest compression ratio. However, it can achieve relatively high compression ratio and keep the highest-fidelity output at the same time. This reveals that \DiffPMAE{}  is ideal for  point cloud stream or data storing tasks by achieving a competitive compression ratio with state-of-the-art decoding performance. In this experiment, we directly use partial point cloud data which can be further compressed by standard libraries. Therefore, the compression gain by \DiffPMAE{} can be further increased. This will be kept as a future work while relating to point cloud streaming tasks.

\begin{table}[]
    \centering
    \begin{adjustbox}{width=0.65\textwidth}
        \begin{tabular}{c|c|c|c|c|c}
        \hline
        Metrics & Draco \cite{draco} & MPEG \cite{octree_3} & G-PCC \cite{gpcc} & D-PCC \cite{dpcc} & Ours \\
        \hline
        \hline
        bpp & 4.51 & 4.29 & 3.10 & \textbf{0.75} & 3.38 \\
        \hline
        CD & 0.0011 & 0.0010 & 0.0010 & 0.0032 & \textbf{0.0004}\\
        \hline
        \end{tabular}
    \end{adjustbox}
    \caption{Comparison of point cloud compression performance.}
    \label{tab:bpp}
\end{table}

\subsection{Ablation study}


\begin{table*}[b]
    \centering
    \small
    \adjustbox{width=\columnwidth,center}{%
    \begin{tabular}{c | c | c | c | c | c | c | c | c | c} 
     \hline
     \multirow{2}{*}{Mask ratio}& \multirow{2}{*}{Evaluation output} & \multicolumn{4}{c|}{Setting~(\textit{a}): Entire point cloud object} & \multicolumn{4}{c}{Setting~(\textit{b}): Only masked regions} \\ 
     \cline{3-10}
     & & MMD CD & 1-NN CD & JSD & HD & MMD CD & 1-NN CD & JSD & HD\\ 
      \hline\hline
 40\% & masked only & 6.187 & 15.869 & 6.636 & 8.902 & 2.373 & 0.244 & 70.119 & 10.950\\ 
 \hline
 50\% & masked only&4.564 & 11.230 & 5.358 & 7.727 & 2.208 & 0.488 & 72.155 & 9.124\\
 \hline
 60\% &masked only &3.720 & 5.126 & 4.935 & 7.035 & 2.052 & 0 & 74.463 & 7.970\\
 \hline
 65\% & masked only&3.420 & 4.150 & 4.610 & 6.750 & 2.018 & 0.244 & 75.927 & 7.496\\
 \hline
 70\% & masked only&3.146 & 3.662 & 4.441 & 6.503 & 1.989 & 0.244 & 76.868 & 7.092\\
 \hline
 75\% &masked only &2.847 & 3.906 & 4.129 & 6.208 & 1.947 & 0.244 & 77.906 & 6.622\\
 \hline
 80\% & masked only&2.660 & 4.882 & 3.960 & 6.031 & 1.944 & 0.976 & 79.337 & 6.347\\
 \hline
 \hline
 75\% & masked+visible& \textbf{1.125} & \textbf{1.464} & \textbf{2.890} & \textbf{3.445} & 1.669 & 18.798 & 50.017 & 4.122\\
 \hline

 80\% & masked+visible & 1.187 & 2.929 & 2.958 & 3.641 & 1.756 & 24.414 & 56.229 & 4.356 \\
 \hline
    \end{tabular}}
    \caption{Ablation study on mask ratio and loss functions. MMD CD, 1-NN CD, JSD results are $\times10^{-3}$, HD results are $\times10^{-2}$. All means combined the predicted masked patches and the visible patches from the GT.
    } 
    \label{tab:mask_ratio_table}
\end{table*}

\noindent
\textbf{Loss function input.}~We  discover that the different loss function settings based on the inputs we use, can create noticeable differences in performance. Thus, in the training phase, we use two different settings for the loss calculation: \textit{a})~Entire point cloud object as the input: we combine the predicted masked parts from the last step of our DM and the input visible parts, then compute loss with initial point cloud, \textit{b})~Only masked region as the input: we directly use the predicted masked parts to compute loss with masked parts from initial point cloud. We have denoted the results under different masked ratios in Table~\ref{tab:mask_ratio_table}. 

Overall, we notice that the best results are achieved under \textit{Setting}~(\textit{a}) when predicting both visible and masked patches by the model (e.g., \textit{Setting}~(\textit{a}) with $r$=75\% \& masked+visible output).  This is because, when training under \textit{Setting}~(\textit{a}), we use the entire object as the ground truth in the loss function, which leads the model to learn distributions more uniformly.
However, such uniform distribution causes \textit{Setting}~(\textit{a}) to perform worse than \textit{Setting}~(\textit{b}) when it is used only the masked parts to assess the performance (e.g., $r=$40--80\% with \textit{Setting}~(\textit{a})). \textit{Setting}~(\textit{b}) is better performed with masked-only assessments, especially when the mask ratio is low due to the availability of more visible patches.

\begin{figure}[t]
  \centering
   \includegraphics[width=0.6\linewidth]{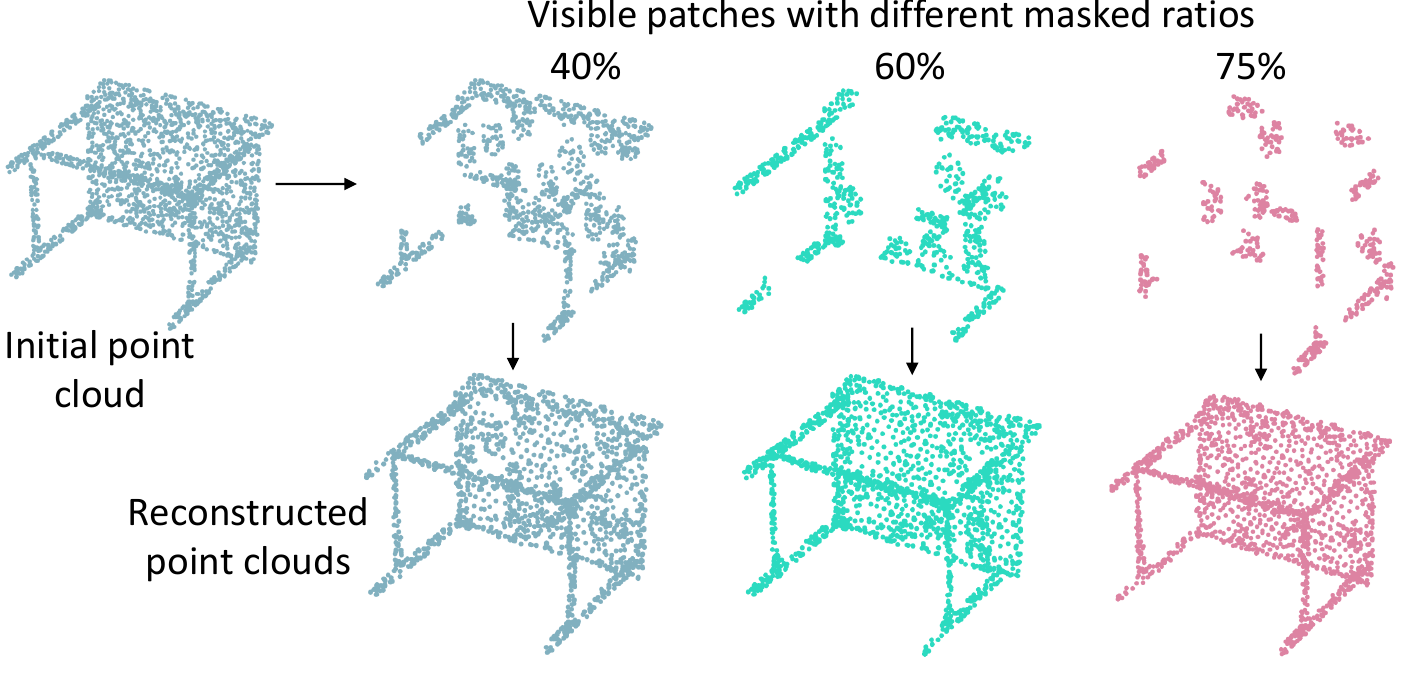}
   \caption{{Visualization of different mask ratios}}
   \label{fig:mask_ratio}
\end{figure}

\noindent
\textbf{Mask ratio ($r$).} 
In default configurations, \DiffPMAE{} decoder predicts only the masked patches. Hence, to avoid the visible patches affecting the performance evaluation, 
we mainly evaluate the generated masked patches only in Table \ref{tab:mask_ratio_table}, which is denoted as \textit{masked only}. 
Interestingly, the optimum mask ratio of our \DiffPMAE{} is 75\%,  which is relatively a higher value. 
Despite the availability of more visible patches, lower $r$ values show higher values in all the metrics considered. 
This is because DM learns features from ground truth of masked patches and uses visible patches as guidance to generate masked patches only. Hence, a lower $r$ reduces the features for the DM to learn and results in decreased performance.
Fig.~\ref{fig:mask_ratio} visualizes predicted output for different mask ratios which emphasizes the same observation where, the higher the mask ratio, the more uniformly distributed and higher quality reconstruction. Further analysis of impact of $r$ on prediction will be done as future work.

\begin{table}[b]
    \centering
    \small{
    \begin{adjustbox}{width=0.65\textwidth}
        \begin{tabular}{c|c|c|c|c}
        \hline
        Strategy & MMD CD {($\times10^{-3}$)} & 1-NN CD {($\times10^{-3}$)} & JSD {($\times10^{-3}$)} & HD {($\times10^{-2}$)}\\
        \hline
        \hline
        Random & 1.125 & 1.464 & 2.890 & 3.445 \\
        \hline
        Block & 1.051 & 0.976 & 3.493 & 3.321 \\
        \hline
        \end{tabular}
    \end{adjustbox}
    }
    \caption{Ablation study on mask strategies.
    }
    \label{tab:mask_str_table}
\end{table}

\begin{figure}
  \centering
   \includegraphics[width=0.7\linewidth]{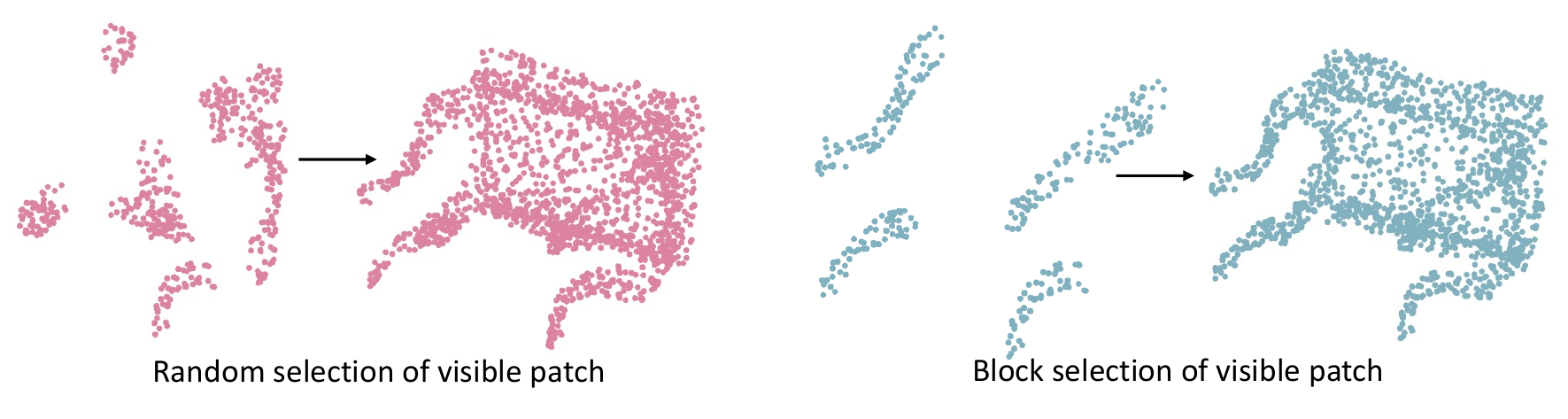}
   \caption{{Different Mask Strategies fro $r=75\%$} The random mask strategy is on the left. The block mask strategy is on the right. 
   }
   \label{fig:mask_str}
\end{figure}

\noindent
\textbf{Mask Strategy.} We now assess the generation performance of our model on different masking strategies. We trained our model with a 75\% mask ratio on two masking strategies: \textit{i})~\textit{random}:  Randomly selecting masked and visible parts based on the given mask ratio after segmentation. \textit{ii})~\textit{block}:  Masking a large block that contains multiple continuous and consecutive patches. We have visualized results with different mask strategies in Fig.~\ref{fig:mask_str}. As we benefit from the robust inference and generation capabilities of our decoder, $D$, we observe no noticeable difference in the performance of our model on block and random strategies. We further verify this observation statistically in  Table \ref{tab:mask_str_table}, where we report less difference in the measured values, particularly in MMD CD and HD values.


\begin{table}[]
    \centering
    \small{
    \begin{adjustbox}{width=0.65\textwidth}
        \begin{tabular}{c|c|c|c|c}
        \hline
        Group & MMD CD ($\times10^{-3}$)  & 1-NN CD ($\times10^{-3}$) & JSD ($\times10^{-3}$) & HD ($\times10^{-2}$) \\
        \hline
        \hline
        G32 N64 & 1.441 & 10.009 & 4.527 & 3.920 \\
        \hline
        G64 N32 & \textbf{1.125} & \textbf{1.464} & \textbf{2.890} & \textbf{3.445} \\
        \hline
        G128 N16 & 1.604 & 18.798 & 192.245 & 3.609 \\
        \hline
        \end{tabular}
    \end{adjustbox}
    }
    \caption{Ablation study on group settings. 
    }
    \label{tab:group_table}
\end{table}

\noindent
\textbf{Group setting.} We observe that number of groups ($G$) and the samples in each group ($N$) has a direct impact on the \DiffPMAE{} performance. Because the K-NN and the FPS algorithms cannot avoid overlapping points, which results in loss of details. Table~\ref{tab:group_table} shows different group configurations along with their measured statistical performance of Decoder output. The best performance is achieved with the configuration $G$64 $N$32 which we consider as a fair balance between the $G$ and $N$. Considering the other settings, $G$32 $N$64 can loss details after segmentation significantly due to the shorter fixed latent length we used. Similarly, for $G$128 $N$16 setting, there are too many patches to process and the latent length is also wasted for 16 points.

\begin{table}[b]
    \centering
    \begin{adjustbox}{width=0.65\textwidth}
        \begin{tabular}{c|c|c|c|c}
        \hline
        Latent Width & MMD CD ($\times10^{-3}$) & 1-NN CD ($\times10^{-3}$) & JSD ($\times10^{-3}$) & HD ($\times10^{-2}$) \\
        \hline
        \hline
        192 & 1.370 & 9.277 & 33.283 & 3.852 \\
        \hline
        384 & 1.125 & 1.464 & \textbf{2.890} & \textbf{3.445} \\
        \hline
        768 & \textbf{1.043} & \textbf{0.488} & 3.576 & 3.356 \\
        \hline
        \end{tabular}
    \end{adjustbox}
    
    \caption{Ablation study on latent width. 
    }
    \label{tab:abl_lw}
\end{table}

\noindent
\textbf{Latent width.}
The width of the latent space represents the amount of detail that the point cloud can retain in the latent space. Therefore, a larger latent space allows the model to capture more features. However, following the increase of the width of latent space, the corresponding resources are also required to increase (i.e., in our experiment, we faced insufficient VRAM when we trained our model with a bigger latent width). To explore that relationship, we discuss the performance of our model with different latent widths in Table \ref{tab:abl_lw}. We use 12 encoder blocks and four decoder blocks; each encoder block contains six heads, and each decoder block contains four heads by default. The mask strategy for this experiment is random, and the mask ratio is 75\%. During the training, we train our encoder for 50 epochs and decoder for 300 epochs. This experiment uses three different latent space widths: 192, 384, and 768. As shown in Table \ref{tab:abl_lw}, our model performs noticeably better when the latent space width is 384 and 768 than the model with 192 latent widths. However, we do not observe significant differences between the width = 384 and 768. That means our model can learn and produce outstanding results by using a limited latent space without excessively increasing the latency width to cause excessive consumption of resources.

\noindent
NOTE: we also evaluate our model on inference speed and diffusion timestep. Please refer to the supplementary for details.

\section{Discussion and Limitation}
\label{sec:Limitation}

\noindent
In the current segmentation process of \DiffPMAE, we observe a slight overlap between the visible and masked patches due to the nature of K-NN+FPS algorithms. This can potentially result in unexpected detail loss of the point cloud data. 
While overlapping-free algorithms like point pairwise are inefficient for complex data, we aim to find more effective approaches in the future. In addition, we aim to apply \DiffPMAE~on more complex point cloud datasets (e.g., LIDAR) to further generalize \DiffPMAE~on different applications. 


\section{Conclusion}
\label{sec:conclusion}

We proposed a novel method, \DiffPMAE{} that leverages a self-supervised learning approach by combining Masked Autoencoding (MAE) and Diffusion Models (DMs) for point cloud reconstruction tasks. The model first converts a given point cloud object to masked and visible regions. Then these visible regions are converted to latent code MAE modules which are taken as conditional input to the DMs for point cloud reconstruction. We further extend \DiffPMAE{} for different point cloud related tasks including point cloud compression, upsampling and completion in which it exceeds the state-of-the-art methods in-terms of quality of reconstruction. In Future work, we aim to further explore the applicability of \DiffPMAE{} for real-time streaming and compression tasks.

\clearpage

%
%
\bibliographystyle{splncs04}
\bibliography{main}

\begin{thebibliography}{10}
\providecommand{\url}[1]{\texttt{#1}}
\providecommand{\urlprefix}{URL }
\providecommand{\doi}[1]{https://doi.org/#1}

\bibitem{draco}
Draco 3d graphics compression. \url{https://google.github.io/draco/}

\bibitem{pcssl_3}
Achituve, I., Maron, H., Chechik, G.: Self-supervised learning for domain adaptation on point clouds. In: Proceedings of the IEEE/CVF winter conference on applications of computer vision. pp. 123--133 (2021)

\bibitem{achlioptas2018learning}
Achlioptas, P., Diamanti, O., Mitliagkas, I., Guibas, L.: Learning representations and generative models for 3d point clouds. In: International conference on machine learning. pp. 40--49. PMLR (2018)

\bibitem{CrossPoint}
Afham, M., Dissanayake, I., Dissanayake, D., Dharmasiri, A., Thilakarathna, K., Rodrigo, R.: Crosspoint: Self-supervised cross-modal contrastive learning for 3d point cloud understanding. In: Proceedings of the IEEE/CVF Conference on Computer Vision and Pattern Recognition (CVPR). pp. 9902--9912 (June 2022)

\bibitem{dae_2}
Bengio, Y., Yao, L., Alain, G., Vincent, P.: Generalized denoising auto-encoders as generative models. In: Proceedings of the 26th International Conference on Neural Information Processing Systems - Volume 1. p. 899–907. NIPS'13, Curran Associates Inc., Red Hook, NY, USA (2013)

\bibitem{hd}
Berger, M., Levine, J.A., Nonato, L.G., Taubin, G., Silva, C.T.: A benchmark for surface reconstruction. ACM Transactions on Graphics (TOG)  \textbf{32}(2),  1--17 (2013)

\bibitem{bbp}
Biswas, S., Liu, J., Wong, K., Wang, S., Urtasun, R.: Muscle: Multi sweep compression of lidar using deep entropy models. Advances in Neural Information Processing Systems  \textbf{33},  22170--22181 (2020)

\bibitem{GPT}
Brown, T., Mann, B., Ryder, N., Subbiah, M., Kaplan, J.D., Dhariwal, P., Neelakantan, A., Shyam, P., Sastry, G., Askell, A., et~al.: Language models are few-shot learners. Advances in neural information processing systems  \textbf{33},  1877--1901 (2020)

\bibitem{shapenet2015}
Chang, A.X., Funkhouser, T., Guibas, L., Hanrahan, P., Huang, Q., Li, Z., Savarese, S., Savva, M., Song, S., Su, H., Xiao, J., Yi, L., Yu, F.: {ShapeNet: An Information-Rich 3D Model Repository}. Tech. Rep. arXiv:1512.03012 [cs.GR], Stanford University --- Princeton University --- Toyota Technological Institute at Chicago (2015)

\bibitem{chen2023introduction}
Chen, A., Mao, S., Li, Z., Xu, M., Zhang, H., Niyato, D., Han, Z.: An introduction to point cloud compression standards. GetMobile: Mobile Computing and Communications  \textbf{27}(1),  11--17 (2023)

\bibitem{p2c}
Cui, R., Qiu, S., Anwar, S., Liu, J., Xing, C., Zhang, J., Barnes, N.: P2c: Self-supervised point cloud completion from single partial clouds (2023)

\bibitem{kdtree_1}
Devillers, O., Gandoin, P.M.: Geometric compression for interactive transmission. In: Proceedings Visualization 2000. VIS 2000 (Cat. No. 00CH37145). pp. 319--326. IEEE (2000)

\bibitem{BERT}
Devlin, J., Chang, M.W., Lee, K., Toutanova, K.: Bert: Pre-training of deep bidirectional transformers for language understanding. arXiv preprint arXiv:1810.04805  (2018)

\bibitem{hrConditionIG_3}
Dhariwal, P., Nichol, A.: Diffusion models beat gans on image synthesis. In: Ranzato, M., Beygelzimer, A., Dauphin, Y., Liang, P., Vaughan, J.W. (eds.) Advances in Neural Information Processing Systems. vol.~34, pp. 8780--8794. Curran Associates, Inc. (2021), \url{https://proceedings.neurips.cc/paper_files/paper/2021/file/49ad23d1ec9fa4bd8d77d02681df5cfa-Paper.pdf}

\bibitem{pcssl_4}
Eckart, B., Yuan, W., Liu, C., Kautz, J.: Self-supervised learning on 3d point clouds by learning discrete generative models. In: Proceedings of the IEEE/CVF conference on computer vision and pattern recognition. pp. 8248--8257 (2021)

\bibitem{cdl2}
Fan, H., Su, H., Guibas, L.J.: A point set generation network for 3d object reconstruction from a single image. In: Proceedings of the IEEE conference on computer vision and pattern recognition. pp. 605--613 (2017)

\bibitem{octree_2}
Golla, T., Klein, R.: Real-time point cloud compression. In: 2015 IEEE/RSJ International Conference on Intelligent Robots and Systems (IROS). pp. 5087--5092. IEEE (2015)

\bibitem{gpcc}
Graziosi, D., Nakagami, O., Kuma, S., Zaghetto, A., Suzuki, T., Tabatabai, A.: An overview of ongoing point cloud compression standardization activities: Video-based (v-pcc) and geometry-based (g-pcc). APSIPA Transactions on Signal and Information Processing  \textbf{9}, ~e13 (2020)

\bibitem{UnsupervisedNLP_3}
Gururangan, S., Marasović, A., Swayamdipta, S., Lo, K., Beltagy, I., Downey, D., Smith, N.A.: Don't stop pretraining: Adapt language models to domains and tasks. In: Proceedings of ACL (2020)

\bibitem{MAE}
He, K., Chen, X., Xie, S., Li, Y., Doll{\'a}r, P., Girshick, R.: Masked autoencoders are scalable vision learners. In: Proceedings of the IEEE/CVF Conference on Computer Vision and Pattern Recognition (2021)

\bibitem{dpcc}
He, Y., Ren, X., Tang, D., Zhang, Y., Xue, X., Fu, Y.: Density-preserving deep point cloud compression. In: Proceedings of the IEEE/CVF Conference on Computer Vision and Pattern Recognition. pp. 2333--2342 (2022)

\bibitem{ddpm}
Ho, J., Jain, A., Abbeel, P.: Denoising diffusion probabilistic models (2020)

\bibitem{compression}
Huang, T., Liu, Y.: 3d point cloud geometry compression on deep learning. In: Proceedings of the 27th ACM international conference on multimedia. pp. 890--898 (2019)

\bibitem{octree_1}
Huang, Y., Peng, J., Kuo, C.C.J., Gopi, M.: A generic scheme for progressive point cloud coding. IEEE Transactions on Visualization and Computer Graphics  \textbf{14}(2),  440--453 (2008)

\bibitem{pfnet}
Huang, Z., Yu, Y., Xu, J., Ni, F., Le, X.: Pf-net: Point fractal network for 3d point cloud completion. In: Proceedings of the IEEE/CVF conference on computer vision and pattern recognition. pp. 7662--7670 (2020)

\bibitem{tetrahedral}
Kalischek, N., Peters, T., Wegner, J.D., Schindler, K.: Tetrahedral diffusion models for 3d shape generation (2022)

\bibitem{kim2023pu}
Kim, D., Shin, M., Paik, J.: Pu-edgeformer: Edge transformer for dense prediction in point cloud upsampling. In: ICASSP 2023-2023 IEEE International Conference on Acoustics, Speech and Signal Processing (ICASSP). pp.~1--5. IEEE (2023)

\bibitem{pugan}
Li, R., Li, X., Fu, C.W., Cohen-Or, D., Heng, P.A.: Pu-gan: a point cloud upsampling adversarial network. In: {IEEE} International Conference on Computer Vision ({ICCV}) (2019)

\bibitem{dispu}
Li, R., Li, X., Heng, P.A., Fu, C.W.: Point cloud upsampling via disentangled refinement. In: Proceedings of the IEEE/CVF conference on computer vision and pattern recognition. pp. 344--353 (2021)

\bibitem{kdtree_2}
Lien, J.M., Kurillo, G., Bajcsy, R.: Multi-camera tele-immersion system with real-time model driven data compression: A new model-based compression method for massive dynamic point data. The Visual Computer  \textbf{26},  3--15 (2010)

\bibitem{pointDiffusion}
Luo, S., Hu, W.: Diffusion probabilistic models for 3d point cloud generation. In: Proceedings of the IEEE/CVF Conference on Computer Vision and Pattern Recognition (CVPR) (June 2021)

\bibitem{octree_3}
Mekuria, R., Blom, K., Cesar, P.: Design, implementation, and evaluation of a point cloud codec for tele-immersive video. IEEE Transactions on Circuits and Systems for Video Technology  \textbf{27}(4),  828--842 (2016)

\bibitem{UnsupervisedNLP_2}
Mikolov, T., Sutskever, I., Chen, K., Corrado, G.S., Dean, J.: Distributed representations of words and phrases and their compositionality. In: Burges, C., Bottou, L., Welling, M., Ghahramani, Z., Weinberger, K. (eds.) Advances in Neural Information Processing Systems. vol.~26. Curran Associates, Inc. (2013), \url{https://proceedings.neurips.cc/paper_files/paper/2013/file/9aa42b31882ec039965f3c4923ce901b-Paper.pdf}

\bibitem{nardo2022point}
Nardo, F., Peressoni, D., Testolina, P., Giordani, M., Zanella, A.: Point cloud compression for efficient data broadcasting: A performance comparison. In: 2022 IEEE Wireless Communications and Networking Conference (WCNC). pp. 2732--2737. IEEE (2022)

\bibitem{tti_2}
Nichol, A., Dhariwal, P., Ramesh, A., Shyam, P., Mishkin, P., McGrew, B., Sutskever, I., Chen, M.: Glide: Towards photorealistic image generation and editing with text-guided diffusion models. arXiv preprint arXiv:2112.10741  (2021)

\bibitem{pointe}
Nichol, A., Jun, H., Dhariwal, P., Mishkin, P., Chen, M.: Point-e: A system for generating 3d point clouds from complex prompts  (Dec 2022)

\bibitem{PointMAE}
Pang, Y., Wang, W., Tay, F.E., Liu, W., Tian, Y., Yuan, L.: Masked autoencoders for point cloud self-supervised learning. In: Computer Vision--ECCV 2022: 17th European Conference, Tel Aviv, Israel, October 23--27, 2022, Proceedings, Part II. pp. 604--621. Springer (2022)

\bibitem{pugcn}
Qian, G., Abualshour, A., Li, G., Thabet, A., Ghanem, B.: Pu-gcn: Point cloud upsampling using graph convolutional networks. In: Proceedings of the IEEE/CVF Conference on Computer Vision and Pattern Recognition. pp. 11683--11692 (2021)

\bibitem{putrans}
Qiu, S., Anwar, S., Barnes, N.: Pu-transformer: Point cloud upsampling transformer. In: Proceedings of the Asian Conference on Computer Vision. pp. 2475--2493 (2022)

\bibitem{quach2022survey}
Quach, M., Pang, J., Tian, D., Valenzise, G., Dufaux, F.: Survey on deep learning-based point cloud compression. Frontiers in Signal Processing  \textbf{2},  846972 (2022)

\bibitem{UnsupervisedNLP_1}
Radford, A., Wu, J., Child, R., Luan, D., Amodei, D., Sutskever, I.: Language models are unsupervised multitask learners  (2019)

\bibitem{tti_1}
Ramesh, A., Dhariwal, P., Nichol, A., Chu, C., Chen, M.: Hierarchical text-conditional image generation with clip latents (2022)

\bibitem{hrConditionIG_2}
Rombach, R., Blattmann, A., Lorenz, D., Esser, P., Ommer, B.: High-resolution image synthesis with latent diffusion models (2021)

\bibitem{pcssl_2}
Sauder, J., Sievers, B.: Self-supervised deep learning on point clouds by reconstructing space. Advances in Neural Information Processing Systems  \textbf{32} (2019)

\bibitem{shi2023enabling}
Shi, Y., Venkatram, P., Ding, Y., Ooi, W.T.: Enabling low bit-rate mpeg v-pcc-encoded volumetric video streaming with 3d sub-sampling. In: Proceedings of the 14th Conference on ACM Multimedia Systems. pp. 108--118 (2023)

\bibitem{hrConditionIG_1}
Sohl-Dickstein, J., Weiss, E., Maheswaranathan, N., Ganguli, S.: Deep unsupervised learning using nonequilibrium thermodynamics. In: Bach, F., Blei, D. (eds.) Proceedings of the 32nd International Conference on Machine Learning. Proceedings of Machine Learning Research, vol.~37, pp. 2256--2265. PMLR, Lille, France (07--09 Jul 2015), \url{https://proceedings.mlr.press/v37/sohl-dickstein15.html}

\bibitem{Song_2023_CVPR}
Song, R., Fu, C., Liu, S., Li, G.: Efficient hierarchical entropy model for learned point cloud compression. In: Proceedings of the IEEE/CVF Conference on Computer Vision and Pattern Recognition (CVPR). pp. 14368--14377 (June 2023)

\bibitem{gecco}
Tyszkiewicz, M.J., Fua, P., Trulls, E.: Gecco: Geometrically-conditioned point diffusion models (2023)

\bibitem{scanobjectnn}
Uy, M.A., Pham, Q.H., Hua, B.S., Nguyen, D.T., Yeung, S.K.: Revisiting point cloud classification: A new benchmark dataset and classification model on real-world data. In: International Conference on Computer Vision (ICCV) (2019)

\bibitem{attention}
Vaswani, A., Shazeer, N., Parmar, N., Uszkoreit, J., Jones, L., Gomez, A.N., Kaiser, {\L}., Polosukhin, I.: Attention is all you need. Advances in neural information processing systems  \textbf{30} (2017)

\bibitem{dae_1}
Vincent, P., Larochelle, H., Bengio, Y., Manzagol, P.A.: Extracting and composing robust features with denoising autoencoders. In: Proceedings of the 25th International Conference on Machine Learning. p. 1096–1103. ICML '08, Association for Computing Machinery, New York, NY, USA (2008). \doi{10.1145/1390156.1390294}, \url{https://doi.org/10.1145/1390156.1390294}

\bibitem{compression_ai_1}
Wang, J., Ding, D., Li, Z., Ma, Z.: Multiscale point cloud geometry compression. In: 2021 Data Compression Conference (DCC). pp. 73--82. IEEE (2021)

\bibitem{cascaded}
Wang, X., Ang~Jr, M.H., Lee, G.H.: Cascaded refinement network for point cloud completion. In: Proceedings of the IEEE/CVF conference on computer vision and pattern recognition. pp. 790--799 (2020)

\bibitem{DiffMAE}
Wei, C., Mangalam, K., Huang, P.Y., Li, Y., Fan, H., Xu, H., Wang, H., Xie, C., Yuille, A., Feichtenhofer, C.: Diffusion models as masked autoencoder. arXiv preprint arXiv:2304.03283  (2023)

\bibitem{modelnet}
Wu, Z., Song, S., Khosla, A., Yu, F., Zhang, L., Tang, X., Xiao, J.: 3d shapenets: A deep representation for volumetric shapes. In: 2015 IEEE Conference on Computer Vision and Pattern Recognition (CVPR). pp. 1912--1920. IEEE Computer Society, Los Alamitos, CA, USA (jun 2015). \doi{10.1109/CVPR.2015.7298801}, \url{https://doi.ieeecomputersociety.org/10.1109/CVPR.2015.7298801}

\bibitem{Snowflake}
Xiang, P., Wen, X., Liu, Y.S., Cao, Y.P., Wan, P., Zheng, W., Han, Z.: Snowflake point deconvolution for point cloud completion and generation with skip-transformer. IEEE Transactions on Pattern Analysis and Machine Intelligence  \textbf{45}(5),  6320--6338 (2023). \doi{10.1109/TPAMI.2022.3217161}

\bibitem{jsd}
Yang, G., Huang, X., Hao, Z., Liu, M.Y., Belongie, S., Hariharan, B.: Pointflow: 3d point cloud generation with continuous normalizing flows. In: Proceedings of the IEEE/CVF international conference on computer vision. pp. 4541--4550 (2019)

\bibitem{foldingnet}
Yang, Y., Feng, C., Shen, Y., Tian, D.: Foldingnet: Point cloud auto-encoder via deep grid deformation. In: Proceedings of the IEEE Conference on Computer Vision and Pattern Recognition (CVPR) (June 2018)

\bibitem{3pu}
Yifan, W., Wu, S., Huang, H., Cohen-Or, D., Sorkine-Hornung, O.: Patch-based progressive 3d point set upsampling. In: Proceedings of the IEEE/CVF Conference on Computer Vision and Pattern Recognition. pp. 5958--5967 (2019)

\bibitem{punet}
Yu, L., Li, X., Fu, C.W., Cohen-Or, D., Heng, P.A.: Pu-net: Point cloud upsampling network. In: Proceedings of the IEEE conference on computer vision and pattern recognition. pp. 2790--2799 (2018)

\bibitem{pointr}
Yu, X., Rao, Y., Wang, Z., Liu, Z., Lu, J., Zhou, J.: Pointr: Diverse point cloud completion with geometry-aware transformers. In: ICCV (2021)

\bibitem{pointbert}
Yu, X., Tang, L., Rao, Y., Huang, T., Zhou, J., Lu, J.: Point-bert: Pre-training 3d point cloud transformers with masked point modeling. In: Proceedings of the IEEE Conference on Computer Vision and Pattern Recognition (CVPR) (2022)

\bibitem{pcn}
Yuan, W., Khot, T., Held, D., Mertz, C., Hebert, M.: Pcn: Point completion network. In: 2018 International Conference on 3D Vision (3DV). pp. 728--737 (2018)

\bibitem{lion}
Zeng, X., Vahdat, A., Williams, F., Gojcic, Z., Litany, O., Fidler, S., Kreis, K.: Lion: Latent point diffusion models for 3d shape generation. In: Koyejo, S., Mohamed, S., Agarwal, A., Belgrave, D., Cho, K., Oh, A. (eds.) Advances in Neural Information Processing Systems. vol.~35, pp. 10021--10039. Curran Associates, Inc. (2022), \url{https://proceedings.neurips.cc/paper_files/paper/2022/file/40e56dabe12095a5fc44a6e4c3835948-Paper-Conference.pdf}

\bibitem{pointm2ae}
Zhang, R., Guo, Z., Gao, P., Fang, R., Zhao, B., Wang, D., Qiao, Y., Li, H.: Point-m2ae: Multi-scale masked autoencoders for hierarchical point cloud pre-training. arXiv preprint arXiv:2205.14401  (2022)

\bibitem{pcssl_1}
Zhang, Z., Girdhar, R., Joulin, A., Misra, I.: Self-supervised pretraining of 3d features on any point-cloud. In: Proceedings of the IEEE/CVF International Conference on Computer Vision. pp. 10252--10263 (2021)

\bibitem{pvd}
Zhou, L., Du, Y., Wu, J.: 3d shape generation and completion through point-voxel diffusion. In: Proceedings of the IEEE/CVF international conference on computer vision. pp. 5826--5835 (2021)

\end{thebibliography}

\title{Supplementary materials for DiffPMAE}
\titlerunning{Supplementary materials for DiffPMAE}
\author{Yanlong Li\inst{1}
\and
Chamara Madarasingha\inst{2} \and
Kanchana Thilakarathna\inst{1}
}
\authorrunning{Y.~Li et al.}
\institute{The University of Sydney\\
\email{yali8838@uni.sydney.edu.au} \email{kanchana.thilakarathna@sydney.edu.au}\\
\and
University of New South Wales\\
\email{c.kattadige@unsw.edu.au}}

\maketitle
\section{Further Evaluations: Real-world scanned dataset}

\begin{figure*}[h]
  \centering
   \includegraphics[width=0.9\linewidth]{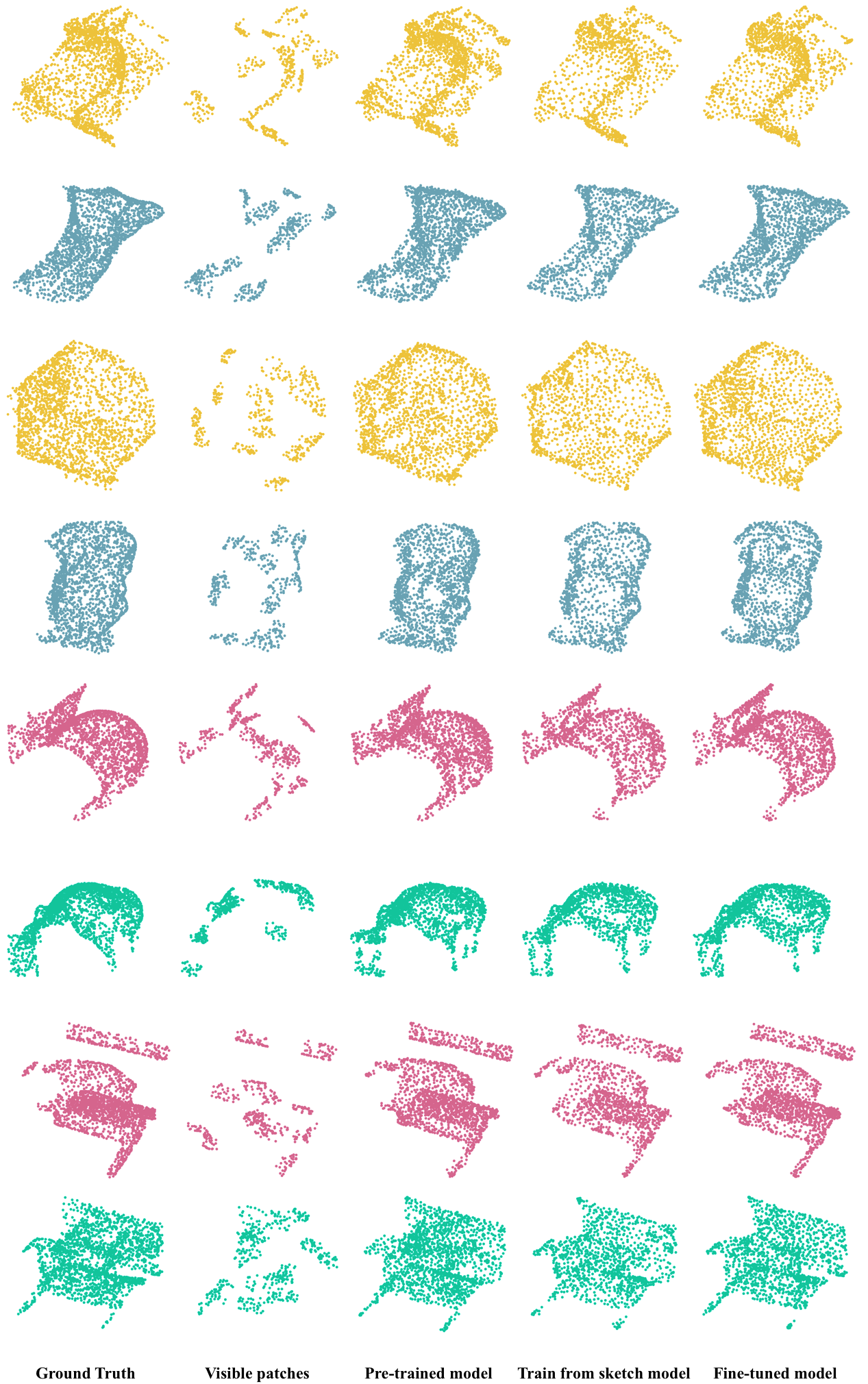}
   \vspace{-2mm}
   \caption{{Reconstruction results of \DiffPMAE{} on ScanObjectNN dataset, main split with background. The predicted results are generated by \DiffPMAE{} with mask ratio $0.75$.}
   }\vspace{-4mm}
   \label{fig:vis_son_w_bg}
\end{figure*}

\begin{figure*}[h]
  \centering
   \includegraphics[width=0.9\linewidth]{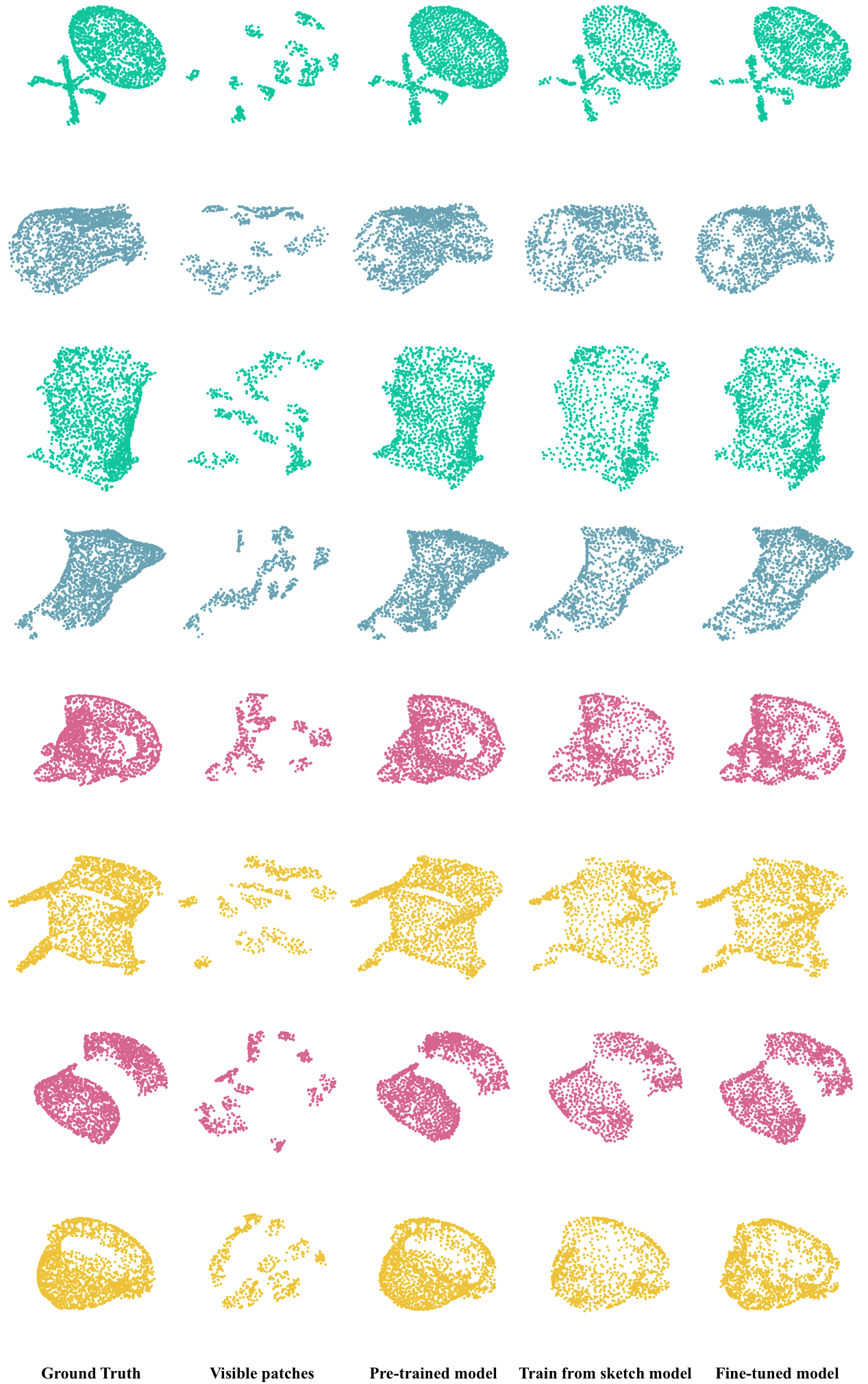}
   \vspace{-2mm}
   \caption{{Reconstruction results of \DiffPMAE{} on ScanObjectNN dataset, main split without background. The predicted results are generated by \DiffPMAE{} with mask ratio $0.75$.}
   }
   \label{fig:vis_son_on_bg}
\end{figure*}

To evaluate the generalization ability of our work, we perform extra experiments on the ScanObjectNN dataset \cite{scanobjectnn}, which contains 2902 unique 3D point cloud objects in 15 categories scanned from real-world objects. We evaluate our model separately on the main split of the ScanObjectNN dataset with background and without background. We take three experiments with different settings of \DiffPMAE{}{} on both with background split and without background split: \textit{i}) \textbf{pre-trained model}, which is the pre-trained on ShapeNet dataset with default configurations (refer to Section 4.1 in our main paper); \textit{ii}) \textbf{train from sketch model}, the \DiffPMAE{} with default configurations trained on ScanObjectNN training set from sketch; \textit{iii}) \textbf{fine-tuned on ScanObjectNN}, the pre-trained \DiffPMAE{} with extra training on ScanObjectNN training set with default configurations. We show the evaluation results with a background in Table \ref{tab:realworld_with_bg}; it reports that the pre-trained \DiffPMAE{} model can achieve competitive generation results even without any fine-tuning on ScanObjectNN dataset. Fig. \ref{fig:vis_son_w_bg} visualizes generated results of three \DiffPMAE{} models. Table \ref{tab:realworld_no_bg} reports the evaluation results without background. Due to objects in this split without background, some objects are incomplete, which is more complex than the main split with the background. Hence, \DiffPMAE{} models perform a bit worse than the split with the background. We visualized the generated results of different \DiffPMAE{} models in Fig. \ref{fig:vis_son_on_bg}. Overall, \DiffPMAE{} can achieve competitive performance on the more complex real-world dataset even without fine-tuning. Those experiments demonstrate the generalization ability of \DiffPMAE{}.

\begin{table}[h]
    \centering
    \adjustbox{width=\columnwidth,center}{
    \begin{tabular}{c|c|c|c}
    \hline
    Models & MMD CD ($\times10^{-3}$) & JSD ($\times10^{-3}$) & HD ($\times10^{-2}$) \\
    \hline
    \hline
    pre-trained model & 1.120  & 5.814 & 3.417 \\
    \hline
    train from sketch model & 1.685  & 119.881 & 4.080 \\
    \hline
    fine-tuned with pre-trained model & 1.244 & 9.505 & 3.535 \\
    \hline
    \end{tabular}}
    \caption{Reconstruction results for \DiffPMAE{} on ScanObjectNN dataset main split with background.
    }
    \label{tab:realworld_with_bg}
\end{table}

\begin{table}[h]
    \centering
    \adjustbox{width=\columnwidth,center}{
    \begin{tabular}{c|c|c|c}
    \hline
    Models & 1-NN CD ($\times10^{-3}$) & JSD ($\times10^{-3}$) & HD ($\times10^{-2}$) \\
    \hline
    \hline
    pre-trained model & 1.131 & 5.934 & 3.355 \\
    \hline
    train from sketch model & 1.300 & 197.363 & 3.486 \\
    \hline
    fine-tuned with pre-trained model & 1.314 & 71.022 & 3.693 \\
    \hline
    \end{tabular}}
    \caption{Reconstruction results for \DiffPMAE{} on ScanObjectNN dataset main split without background.
    }
    \label{tab:realworld_no_bg}
\end{table}

\section{Further Evaluation: Validation of Completion on PCN dataset}

\begin{table}[]
    \centering
    \small
    \begin{tabular}{c|c|c|c|c}
    \hline
    Metrics & PCN & SnowFlake & P2C \cite{p2c} & Ours \\
    \hline
    \hline
    MMD CD & 5.22 & 2.32 & 1.22 & 1.49 \\
    \hline
    \end{tabular}
    \caption{Comparison on PCN dataset. MMD CD results are $\times10^{-3}$.}
    \label{tab:pcn_set}
\end{table}

In Table. \ref{tab:pcn_set}, we also show the comparison with other models on PCN dataset\cite{pcn}. Performance of DiffPMAE and P2C are close but P2C is slightly better than ours. We emphasize that completion is only one of our downstream tasks, therefore, with further fine-tuning, DiffPMAE could achieve further improvements for completion, which we keep in future work.

\section{Further Evaluation: Inference speed}

Our model can produce around 4 items/sec on a one RTX3080 with the default setting. We tested the inference speed with different $t$ settings in Table \ref{tab:inf_spd}. To allocate the computational resource effectively, we also test the parallel sampling with default setting ($t=200$) by setting the batch size to 32. With that setting, our model can produce 31 items/sec.  

\begin{table}[h]
    \centering
    \adjustbox{width=\columnwidth,center}{
    \begin{tabular}{c|c|c|c|c}
    \hline
     & $t=100$, batch size = 1 & $t=200$, batch size = 1 & $t=300$, batch size = 1 &      $t=200$, batch size = 32\\
    \hline
    \hline
    items/sec & 8.22  & 4.08 & 1.89 & 31.03 \\
    \hline
    \end{tabular}}
    \caption{Inference speed for different t settings.
    }
    \label{tab:inf_spd}
\end{table}
\section{Further Ablation: Diffusion Timestep}

To measure the impact of timestemp $t$ on diffusion process, we set timestep $t$=50, 100, 200, 300. We kept other hyperparameters as default as in Section 4.1 in our main paper. According to Table~\ref{tab:abl_diff_time}, \DiffPMAE{} performs best at $t=200$. However, further increase in $t$ reduces the quality of the process in all metrics considered.
The main reason is that with the increase of $t$, the required number of epochs should also be increased which is not scalable. Hence, we select $t=200$ in \DiffPMAE{} which significantly reduces time consumption compared to other point cloud generation methods~ \cite{pointe, lion, tetrahedral} that requires $t$ around 1000.

\begin{table}[]
    \centering
    \begin{adjustbox}{width=0.65\textwidth}
        \begin{tabular}{c|c|c|c|c}
        \hline
        $t$ & MMD CD ($\times10^{-3}$) & 1-NN CD ($\times10^{-3}$) & JSD ($\times10^{-3}$) & HD ($\times10^{-2}$) \\
        \hline
        \hline
        $t=50$ & 1.474 & 11.718 & 4.527 & 3.891 \\
        \hline
        $t=100$ & 1.449 & 12.939 & 117.491 & 3.713 \\
        \hline
        $t=200$ & \textbf{1.125} & \textbf{1.464} & \textbf{2.890} & \textbf{3.445} \\
        \hline
        $t=300$ & 1.506 & 8.544 & 71.055 & 3.842 \\
        \hline
        \end{tabular}
    \end{adjustbox}
    
    \caption{Impact of diffusion timestep.
    }
    \label{tab:abl_diff_time}
\end{table}

\section{Further Ablation: Validation of Diffusion Model Parameters}

We perform further ablation study with 300 diffusion steps with different parameters alongside the diffusion timestep experiments in our main paper. We show results with different setups in Table \ref{tab:abl_diff_time}. We first extend the training process by increasing 300 epochs to 600 epochs. Then, we test a more complex decoder configuration with 8 depths and 6 headers. We also test the different $\Delta T$ values that control the diffusion process. However, none of those configurations with $t=300$ can beat the default configuration of our model with $t=200$. Those experiments further prove that our work can achieve better results with a more simple structure with $t=200,$ and those experiments also prove our method is efficient. Compared with prior diffusion-based works like PVD \cite{pvd}, our work only requires 200 steps to generate high-fidelity results instead of 1000 steps.

\label{sec:more_ablation}
\begin{table}
    \centering
    \adjustbox{width=\columnwidth,center}{
    \begin{tabular}{c|c|c|c|c}
    \hline
    Model configurations & MMD CD ($\times10^{-3}$) & 1-NN CD ($\times10^{-3}$) & JSD ($\times10^{-3}$) & HD ($\times10^{-2}$) \\
    \hline
    \hline
    $t=200$ & \textbf{1.125} & \textbf{1.464} & \textbf{2.890} & \textbf{3.445} \\
    \hline
    $t=300$ & 1.506 & 8.544 & 71.055 & 3.842 \\
    \hline
    $t=300 (e=600)$ & 2.128 & 46.386 & 138.907 & 4.891 \\
    \hline
    $t=300 (d=8, h=6, e=600)$ & 1.792 & 26.367 & 64.276 & 4.427 \\
    \hline
    $t=300 (d=8, h=6, e=600, \Delta T=0.02)$ & 1.653 & 15.869 & 22.556 & 4.251 \\
    \hline
    \end{tabular}}
    \caption{Impact of different model configurations for \DiffPMAE{}. $t$ is number of diffusion steps, $e$ is number of training epochs, $d$ is number of depth in decoder, $h$ is number of header in decoder.
    }
    \label{tab:abl_diff_time}
\end{table}

\vspace{-4mm}
\section{Qualitative Results: Visualizations}

\begin{figure*}[h]
  \centering
   \includegraphics[width=\linewidth]{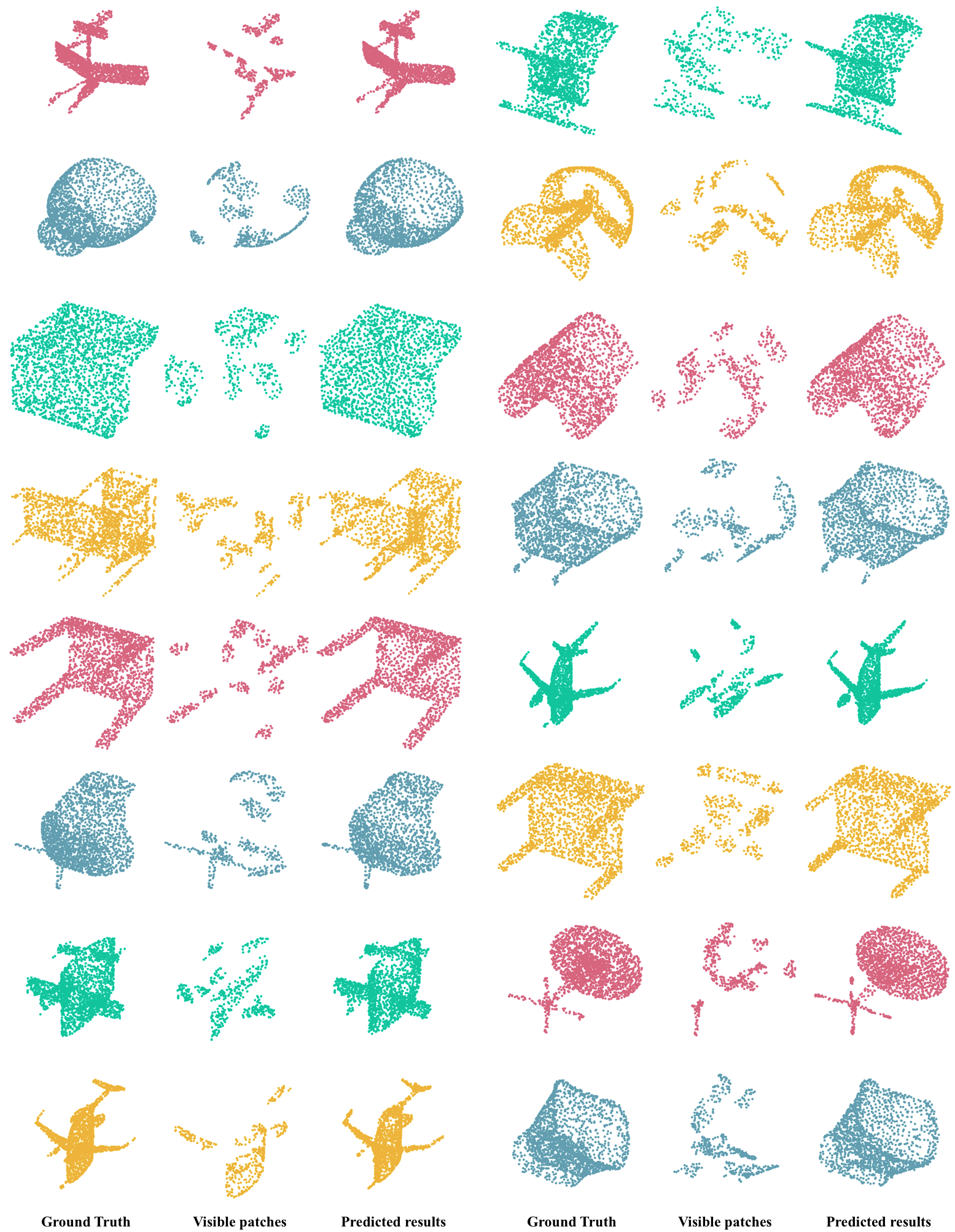}
   \vspace{-2mm}
   \caption{{Reconstruction results of \DiffPMAE{} with multiple categories. Input point cloud and predicted results are 2048 points. The predicted results are generated by \DiffPMAE{} with mask ratio $0.75$.}
   }
   \label{fig:vis_recons}
\end{figure*}

\begin{figure*}[h]
  \centering
   \includegraphics[width=0.7\linewidth]{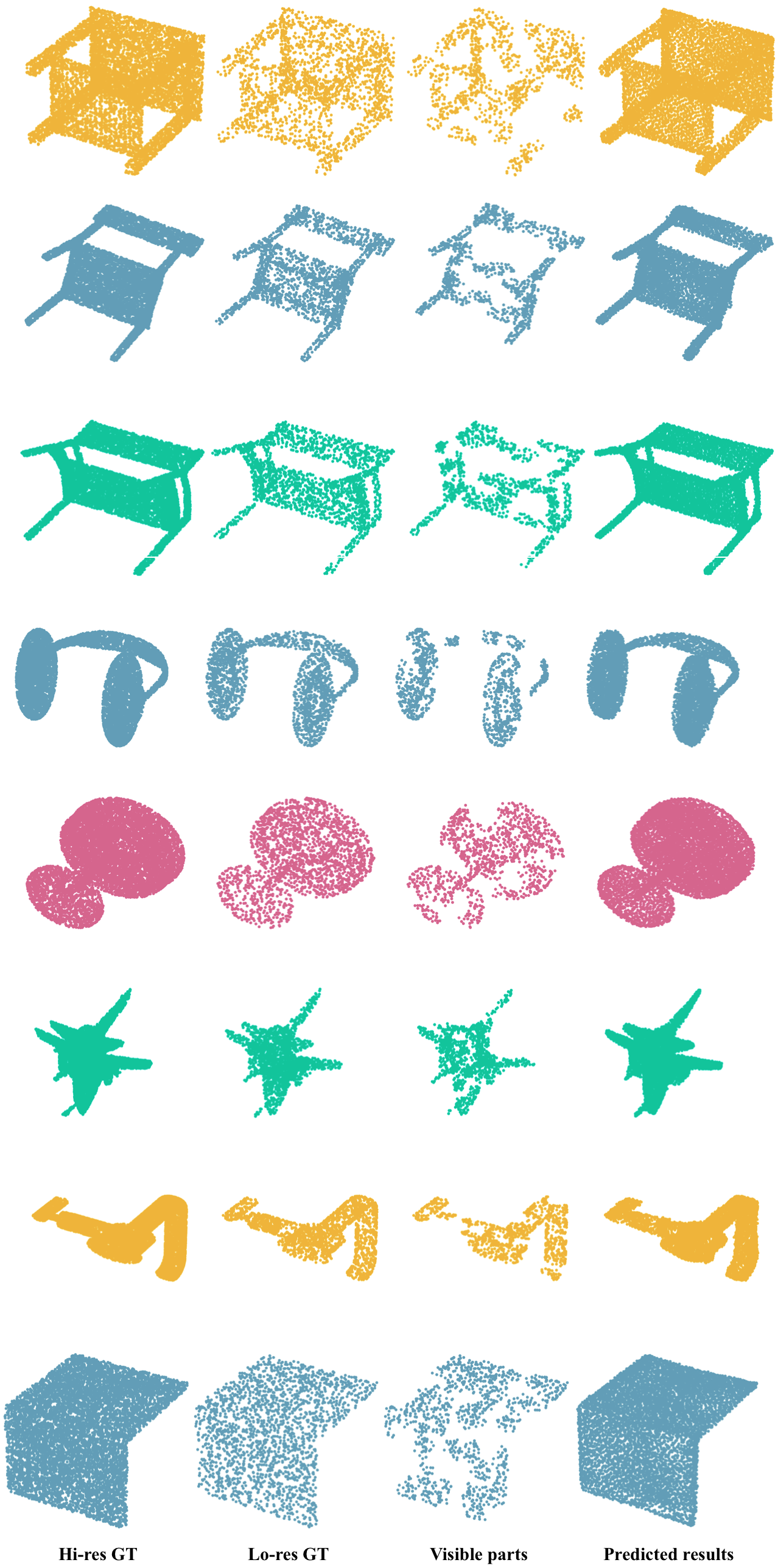}
   \vspace{-2mm}
   \caption{{Upsampling results of \DiffPMAE{} with multiple categories. Input low resolution point clouds are contains 2048 points, high resolution and generated results are contains 8192 points. Our model is trained based on pairs of Hi-res GT and Lo-res GT and use Visible parts to generate the Hi-res predicted results. The predicted results are generated by \DiffPMAE{} with mask ratio $0.4$.}
   }
   \label{fig:vis_sr}
\end{figure*}

In this section, we present further evidence for the quality of results for reconstruction and upsampling tasks that were presented in the main paper. For reconstruction tasks, we use the trained \DiffPMAE{} with the default configuration and parameters in our main paper. For upsampling tasks, we use the trained \DiffPMAE{} for upsampling correspondingly (See Section 4.3 Upsampling part in our main paper). Fig. \ref{fig:vis_recons} shows the reconstruction results of \DiffPMAE{}, with both input ground truth and output predicted results of 2048 points. Fig. \ref{fig:vis_sr} shows the upsampling results of \DiffPMAE{}; the input size is 2048, and the output size is 8192.
\definecolor{codegreen}{rgb}{0,0.6,0}
\definecolor{codegray}{rgb}{0.5,0.5,0.5}
\definecolor{codepurple}{rgb}{0.58,0,0.82}
\definecolor{backcolour}{rgb}{0.95,0.95,0.95}

\lstdefinestyle{python}{
  backgroundcolor=\color{backcolour},   commentstyle=\color{codegreen},
  keywordstyle=\color{magenta},
  numberstyle=\tiny\color{codegray},
  stringstyle=\color{codepurple},
  basicstyle=\ttfamily\tiny,
  breakatwhitespace=true,         
  breaklines=true,                 
  captionpos=b,                    
  keepspaces=true,                 
  numbers=left,                    
  numbersep=5pt,                  
  showspaces=false,                
  showstringspaces=false,
  showtabs=false,                  
  tabsize=5
}
\lstset{style=python}

\section{Diffusion code}

In this section, we provide a simple diffusion code of \DiffPMAE{}. The full code repo can be find in our main paper.

\begin{lstlisting}[language=Python]
class Diff_Point_MAE(nn.Module):
    def sampling_t(self, noisy_t, t, mask, center, x_vis):
        """
        Reverse sampling at timestep t.
        Input noisy level at timestep t,
        return noisy level at timestep t-1.
        """
        B, _, C = x_vis.shape  # B VIS C
        ts = self.time_emb(t.to(x_vis.device)).unsqueeze(1).expand(-1, self.num_group, -1)
        betas_t = self.get_index_from_list(self.betas, t, noisy_t.shape).to(x_vis.device)

        pos_emd_vis = self.decoder_pos_embed(center[~mask]).reshape(B, -1, C)
        pos_emd_msk = self.decoder_pos_embed(center[mask]).reshape(B, -1, C)
        pos_full = torch.cat([pos_emd_vis, pos_emd_msk], dim=1)
        _, N, _ = pos_emd_msk.shape
        mask_token = self.mask_token(noisy_t.reshape(B, N, -1).transpose(1, 2)).transpose(1, 2).to(x_vis.device)
        x_full = torch.cat([x_vis, mask_token], dim=1)
        x_rec = self.MAE_decoder(x_full, pos_full, N, ts)
        x_rec = self.increase_dim(x_rec.transpose(1, 2)).transpose(1, 2).reshape(B, -1, 3)

        alpha_bar_t = self.get_index_from_list(self.alpha_bar, t, noisy_t.shape).to(x_vis.device)
        alpha_bar_t_minus_one = self.get_index_from_list(self.alpha_bar_t_minus_one, t, noisy_t.shape).to(x_vis.device)
        sqrt_alpha_t = self.get_index_from_list(self.sqrt_alphas, t, noisy_t.shape).to(x_vis.device)
        sqrt_alphas_bar_t_minus_one = self.get_index_from_list(self.sqrt_alpha_bar_minus_one, t, noisy_t.shape).to(
            x_vis.device)

        model_mean = (sqrt_alpha_t * (1 - alpha_bar_t_minus_one)) / (1 - alpha_bar_t) * noisy_t + (
                    sqrt_alphas_bar_t_minus_one * betas_t) / (1 - alpha_bar_t) * x_rec

        sigma_t = self.get_index_from_list(self.sigma, t, noisy_t.shape).to(x_vis.device)

        if t == 0:
            return model_mean
        else:
            return model_mean + torch.sqrt(sigma_t) * x_rec

    def sampling(self, x_vis, mask, center, trace=False, noise_patch=None):
        """
        Sampling the masked patches from Gaussian noise.
        """
        B, M, C = x_vis.shape
        if noise_patch is None:
            noise_patch = torch.randn((B, (self.num_group - M) * self.group_size, 3)).to(x_vis.device)
        diffusion_sequence = []

        for i in range(0, self.timestep)[::-1]:
            t = torch.full((1,), i, device=x_vis.device)
            noise_patch = self.sampling_t(noise_patch, t, mask, center, x_vis)
            if trace:
                diffusion_sequence.append(noise_patch.reshape(B, -1, 3))

        if trace:
            return diffusion_sequence
        else:
            return noise_patch.reshape(B, -1, 3)

\end{lstlisting}

\end{document}